\documentclass[10pt,twocolumn,letterpaper]{article}

\usepackage{cvpr}              % To produce the CAMERA-READY version
\usepackage[accsupp]{axessibility}  % Improves PDF readability for those with disabilities.

\usepackage{multirow}

\usepackage{url}
\usepackage{booktabs}
\usepackage{graphicx}
\usepackage{xcolor}
\usepackage{multirow}
\usepackage{colortbl}
\usepackage{subcaption}
\usepackage{siunitx}
\usepackage{amssymb}
\usepackage{tabularx}
\usepackage[ruled]{algorithm2e}

\definecolor{best}{rgb}{0.96, 0.57, 0.58}
\definecolor{second}{rgb}{0.98, 0.78, 0.57}
\definecolor{third}{rgb}{1.0, 1.0, 0.56}

\definecolor{cvprblue}{rgb}{0.21,0.49,0.74}
\usepackage[pagebackref,breaklinks,colorlinks,allcolors=cvprblue]{hyperref}

\def\paperID{6750} % *** Enter the Paper ID here
\def\confName{CVPR}
\def\confYear{2025}

\title{GS-2DGS: Geometrically Supervised 2DGS for Reflective Object Reconstruction}
\author{Jinguang Tong$^{1,2}$, Xuesong Li$^{1,2}$, Fahira Afzal Maken$^{2}$, Sundaram Muthu$^{2,3}$ \\ Lars Petersson$^{2}$, Chuong Nguyen$^{1,2}$, Hongdong Li$^{1}$ \\ 
{\small $^{1}$Australian National University \quad $^{2}$CSIRO \quad $^{3} $Indian Institute of Technology Madras}
}

\begin{document}
\maketitle
\begin{abstract}

% \TODO{}{Complete abstract.}

% 3D modeling of highly reflective objects remains a challenging task due to strong view-dependent appearance. % and smooth textureless surfaces. 
% Previous SDF-based methods can recover high-quality meshes, but these methods are quite time-consuming and tend to produce over-smoothed surfaces. 3D Gaussian Splatting (3DGS) has shown the advantage of high speed and high-detailed rendering in real-time,  but surface extraction from the Gaussians tends to be noisy due to lack of geometry constraint. To bridge the gap between these approaches, we propose a novel reconstruction method for reflective objects that is based on 2D Gaussian splatting (2DGS). This approach combines the high rendering speed of Gaussian Splatting and gains additional geometric information from a foundation model. Experiment results on both the synthetic and real datasets show our method surpass Gaussian-based methods by a significant margin and are very comparable to SDF-based methods while being an order of magnitude faster
3D modeling of highly reflective objects remains challenging due to strong view-dependent appearances. While previous SDF-based methods can recover high-quality meshes, they are often time-consuming and tend to produce over-smoothed surfaces. In contrast, 3D Gaussian Splatting (3DGS) offers the advantage of high speed and detailed real-time rendering, but extracting surfaces from the Gaussians can be noisy due to the lack of geometric constraints. To bridge the gap between these approaches, we propose a novel reconstruction method called GS-2DGS for reflective objects based on 2D Gaussian Splatting (2DGS). Our approach combines the rapid rendering capabilities of Gaussian Splatting with additional geometric information from foundation models. Experimental results on synthetic and real datasets demonstrate that our method significantly outperforms Gaussian-based techniques in terms of reconstruction and relighting and achieves performance comparable to SDF-based methods while being an order of magnitude faster. Code is available at \url{https://github.com/hirotong/GS2DGS}
% By leveraging the additional supervision from foundation models, our method is able to produce much more accurate reconstruction than other Gaussian-based methods.    %for high rendering speed and surface normal and depth are generated from a foundation model to reduce noise.}
% Our method achieves state-of-the-art performance, surpassing Gaussian Splatting-based methods by a significant margin, and it achieves results on-par with SDF based methods while being an order of magnitude faster.
\end{abstract}
\section{Introduction}
\label{sec:intro}

% \TODO{}{The introduction need proofreading and polish the language.}

Modeling 3D assets of specular objects from multi-view images has been a long-standing and challenging task in computer graphics and vision, since specular reflections are view-dependent and thus violate the multi-view consistency assumption adopted by most reconstruction methods. Previous works \cite{liu2023nero, li2024tensosdf} achieve promising results by using a neural radiance field (NeRF) and a signed distance field (SDF). However, these methods require a significant computing effort to train, sometimes hours for a single scene.

The difficulty of reconstructing reflective objects lies in the ill-posed nature of the problem. The appearance of a reflective surface is determined by a combination of the surface properties (material and geometry) and the lighting condition of the environment~\cite{kajiya1986rendering}, while previous methods \cite{liu2023nero, li2024tensosdf} consider only one of these factors.

Since 3D Gaussian Splatting (3DGS) \cite{kerbl20233dgs} achieves significantly higher rendering speed with greater detail and photo consistency, it has been widely used in 3D scene modeling.
%On the aspect of geometry reconstruction, 
However, the surface reconstruction of 3DGS needs further improvement. On the one hand, several Gaussian-based methods \cite{dai2024high,huang20242dgs,yu2024gsdf} attempt to improve reconstruction quality by imposing additional constraints on the surface geometry by flattening the Gaussians or incorporating SDF for additional smoothness. Such approaches achieve superior reconstruction quality on common multi-view images with a much improved computational efficiency.
On the other hand, some other works \cite{jiang2024gaussianshader,liang2024gs,gao2023relightable} address the rendering performance and enable relighting of Gaussian-based methods by introducing physical-based rendering and inverse rendering (IR) techniques. %However, these methods still have limitations in reconstructing reflective objects.
It should be noted, however, that these methods still have limitations in reconstructing reflective objects as they only consider one aspect of the problem.
% On the one hand, recent advances in Gaussian-based methods \cite{dai2024high,huang20242d,yu2024gsdf} tried to improve reconstruction quality by imposing additional constraints on the surface geometry by flatten Gaussian or   rendering quality and reconstruction quality on common multiview images with much higher time efficiency. Some following works \cite{jiang2024gaussianshader,liang2024gs,gao2023relightable} tried to improve the relighting performance of Gaussian-based methods by introducing physical-based rendering and inverse rendering (IR) techniques. However, these methods still have limitations in reconstructing reflective objects.

A naive approach of simply combining these two aspects is not enough to address the issue of reflective object reconstruction, because of the ill-posed nature of the problem. To this end, a straightforward approach is to include additional constraints either on the properties of the surface or the lighting condition.
Recent works \cite{bochkovskii2024depth,ke2024repurposing,fu2025geowizard,martingarcia2024diffusione2eft} based on transformer or diffusion models and trained with a large amount of real and synthetic data have shown impressive performance on monocular geometry (depth and normal) estimation. Unlike multiview stereo methods, which infer geometry information from multiview consistency, these methods directly predict geometry from a single view based on the knowledge of massive training data, thus making them insusceptible to the reflective surface problem.

In this paper, we propose a novel framework for modeling reflective object reconstruction by leveraging foundation models in physical-based Gaussian splatting. By incorporating additional information on the surface, our method can simultaneously address the issues arising from both surface-related properties (geometry and material) and lighting conditions. Our method is achieving state-of-the-art results by some margin with respect to Gaussian Splatting based methods, and it is achieving results on par with NeRF-based methods while being an order of magnitude faster.

\noindent To summarize, our contributions are as follows:
\begin{itemize}
    \item We propose a novel framework for reflective object reconstruction by leveraging foundation models in physical-based Gaussian splatting.
    \item We introduce the deferred shading technique to improve the estimation of environmental lighting.
    \item We achieve state-of-the-art results on reflective object reconstruction by addressing both surface-related properties and lighting conditions.
\end{itemize}

\section{Related work}
\label{sec:related_work}

% \TODO{Sundaram or Chuong}{Need help to write this part. Neural geometry reconstruction and Reflective object reconstruction can be put together. }

\subsection{Radiance field for geometry reconstruction}

Neural reconstruction methods can be divided into the following categories: Neural Radiance Fields (NeRF), Signed Distance Fields (SDF) and Gaussian Splatting (GS). First, NeRF \cite{mildenhall2021nerf} is well known as a pioneering approach for novel view synthesis of complex scenes based on Multi-Layer Perception (MLP). Variants of NeRF include Mip-NeRF \cite{barron2021mip} for antialiasing, Plenoctrees \cite{yu2021plenoctrees} for fast rendering, and Instant NeRF~\cite{zhao2023instant} for fast reconstruction. Second, SDF approach is popularized by NeuS \cite{zhao2023instant} where an MLP is used to encode SDF to model the 3D surface of the scene. Recent SDF approaches include NeuS2\cite{wang2023neus2} for fast neural surface reconstruction using multi-resolution hash encodings, Neuralangelo \cite{li2023neuralangelo} for using higher-order derivatives supervision. Finally, GS was recently proposed by \cite{kerbl20233dgs} as a fast and high photo consistency alternative for neural rendering.  Further extensions include 2DGS \cite{huang20242dgs} for using 2D Gausians instead of 3D Gaussian, PGSR \cite{chen2024pgsr} for flattening 3D Gaussians along their smallest axis, and GSDF \cite{yu2024gsdf} for combining the best of SDF and GS.

\subsection{Reflective object reconstruction}

Reconstructing highly reflective objects is challenging and requires special treatment \cite{ihrke2010transparent, guo2022nerfren, verbin2022ref, tong2023seeing}. Traditional approaches for reconstructing transparent and reflective objects as reported by \cite{ihrke2010transparent} require deep knowledge of optical phenomena and physical modeling that treat specific cases separately. Recent neural radiance-based approaches allow a range of complex optical phenomena and larger scenes to be modeled more effectively.

NeRF-based approaches such as NeRFReN \cite{guo2022nerfren} model the reflection by including a second MLP branch, while Ref-NeRF \cite{verbin2022ref} replaces NeRF's view dependence with Integrated Directional Encoding, and Planar Reflection-Aware NeRF \cite{gao2024planar} identifies and models planar reflectors as a ray tracing problem.

For SDF-based approaches,  NeRO \cite{liu2023nero} includes a novel light representation based on slit-sum approximation for highly reflective objects, while TensorSDF \cite{li2024tensosdf} employs a surface roughness-aware tensorial representation. 

For GS approaches, R3DG~\cite{gao2023relightable} proposes point-based ray-tracing to improve the relighting quality of reflective objects, while GS-IR~\cite{liang2024gs} and GaussianShader~\cite{jiang2024gaussianshader} focus on incorporating inverse rendering and simplified approximation of the rendering equation, respectively. 

% \subsection{Inverse Rendering}

% Neural-PBIR \cite{sun2023neural} reconstruction of shape, material, and illumination, NeRO \cite{liu2023nero}: Neural Geometry and BRDF Reconstruction of Reflective Objects from Multiview Images

\subsection{Monocular depth/normal estimation}

Monocular depth estimation involves predicting depth from a single RGB image~\cite{saxena2008make3d,eigen2014depth}, a task that has progressed from multi-scale networks~\cite{eigen2015predicting} to advanced approaches like transformers~\cite{yang2021transformer, zhao2022monovit,ranftl2021vision} and diffusion models~\cite{saxena2024surprising,mao2024stealing}. Methods such as MiDaS~\cite{ranftl2020towards} and MegaDepth~\cite{li2018megadepth} leverage large-scale datasets to achieve affine-invariant depth predictions, which generalize well to unseen scenes but remain ambiguous in scale and shift, limiting their applicability for metric depth estimation. Diffusion-based models, including DiffusionDepth~\cite{duan2025diffusiondepth} and VPD~\cite{zhao2023unleashing}, have further improved cross-domain generalization by leveraging pre-trained latent diffusion models. However, these methods often require fine-tuning or additional inputs, such as camera intrinsics, making them computationally demanding.

Surface normal estimation~\cite{zhao2023unleashing}, on the other hand, captures local geometric details without metric ambiguities, making it valuable for tasks like scene reconstruction and object localization. Methods like GeoNet~\cite{yin2018geonet} and DSINE~\cite{bae2024rethinking} enhance generalization by enforcing depth-normal consistency and leveraging diverse datasets, while OmniData~\cite{eftekhar2021omnidata} provides large-scale annotated normals for improved training.
These approaches aim to improve generalization, sharpen occlusion boundaries, and reduce dependency on extensive labeled datasets, offering a pathway to more robust and scalable solutions. In this work, we propose to leverage these advancements to improve the reconstruction of reflective objects from monocular images.

% Recent works show promise in integrating diffusion priors and self-supervised learning to enhance both depth and normal estimation~\cite{mao2024stealing}. 

% These approaches aim to improve generalization, sharpen occlusion boundaries, and reduce dependency on extensive labeled datasets, offering a pathway to more robust and scalable solutions.Recent works show promise in integrating diffusion priors and self-supervised learning to enhance both depth and normal estimation~\cite{mao2024stealing}. 
\section{Method}
\label{sec:method}

\begin{figure*}[t]
    \centering
    \includegraphics[width=1.0\linewidth]{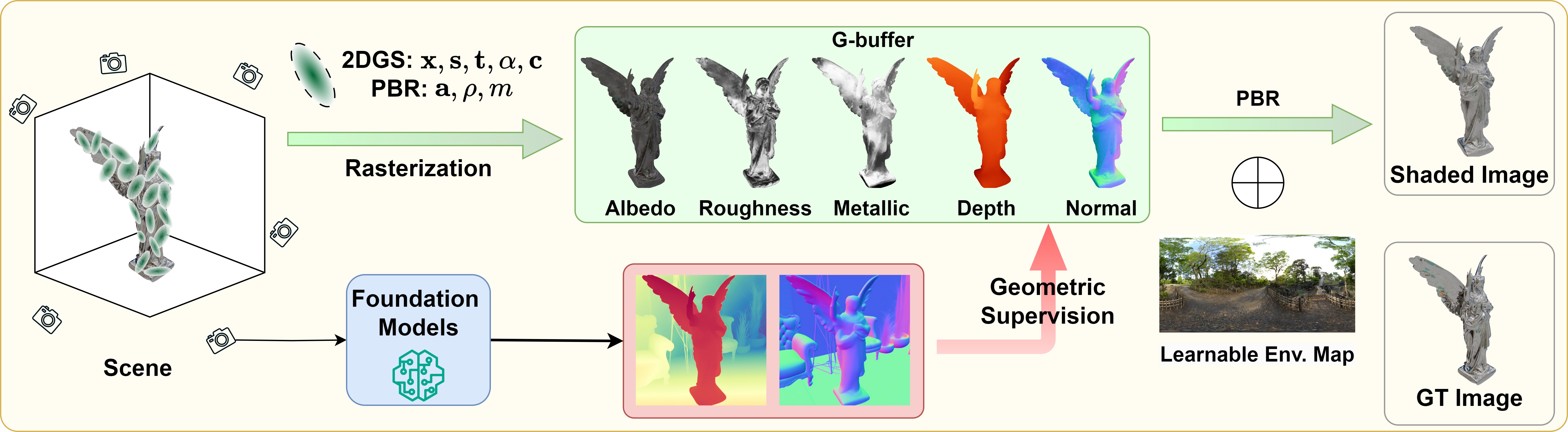}
    \caption{The overview of our GS-2DSG method for reflective object reconstruction. The foundation models supervise 2DGS reconstruction via estimated depth and normal maps to improve geometry and surface smoothness. An environment map is also learned and rendered to account for reflections.}
    \label{fig:method_overview}
\end{figure*}

\subsection{Preliminary: 2D Gaussian Splatting}
\label{sec:method_preliminary}

Since we focus on achieving high-quality reconstruction, our method builds upon the state-of-the-art surfel-based 2DGS~\cite{huang20242dgs} due to its superior geometry performance and efficiency. 2DGS proposes to collapse the 3D volume into a set of 2D-oriented planar Gaussian disks and introduces a perspective-accurate 2D splatting process. Similar to 3DGS, the 2D splat is characterized by its central point \(p_k\), two principal tangential vectors \(t_u\) and \(t_v\), and a corresponding scaling vector \(S = (s_u, s_v)\) that controls the variances of the 2D Gaussian distribution. Compared to 3DGS, 2DGS represents the scene's geometry better because the oriented planar Gaussian can be perfectly aligned to the surface and the normal direction is well-defined as the normal of the plane.

To be specific, a 2D Gaussian disk is defined in a local tangent space in world space, parameterized as:
\begin{equation}
    P(u, v) = \mathbf{x} + s_{u}\mathbf{t}_{u}u + s_v\mathbf{t}_{v}v
\end{equation}
%where 
And for the point \(\mathbf{u} (u, v)\) in the \(uv\) space, its 2D Gaussian value can then be calculated by standard Gaussian
\begin{equation}
    \mathcal{G}(\mathbf{u})=\exp\left(-\frac{u^2+v^2}2\right)
\end{equation}
The center \(\mathbf{x}\), scaling \((s_u, s_v)\), and the rotation \((\mathbf{t}_u, \mathbf{t}_v)\) are learnable parameters. Following 3DGS \citep{kerbl20233dgs}, each 2D Gaussian primitive has opacity \(\alpha\) and view-dependent appearance \(\mathbf{c}\) parameterized with spherical harmonics.

Instead of projecting the 2D Gaussian primitives onto the image space for rendering, 2DGS derives the intersection point of ray and splat in the local tangent space by plane intersection, which alleviates the problem of splat degeneration, especially at grazing angle. The rasterization process is similar to 3DGS, in that 2D Gaussians are sorted based on the depth of their center, and volumetric alpha blending is used to integrate alpha-weighted appearance from front to back:
\begin{equation}\label{eq:color_integration}
    \mathbf{c}(\mathbf{x})=\sum_{i=1}\mathbf{c}_i \alpha_i \hat{G}_i(\mathbf{u}(\mathbf{x}))\prod_{j=1}^{i-1}(1-\alpha_j \hat{G}_j(\mathbf{u}(\mathbf{x})))
\end{equation}
where \(\hat{\mathcal{G}}\) is the object-space low-pass filter introduced in \citep{botsch2005high}. Please refer to the original paper for more details.

\subsection{Geometric properties from Gaussian}

To reconstruct the surface of the object and utilize the geometric information from the foundation models, we derive the depth and normal rendering from the 2D Gaussian primitives.
\paragraph{Normal} Different from other 3DGS-based methods \citep{chen2023neusg,chen2024vcr} that assume the direction of the axis with a minimum scale factor as the normal direction, the 2D Gaussian primitives in 2DGS have a well-defined normal direction as the normal of the plane. The normal of each 2D Gaussian primitive can be calculated by the cross product of the two principal tangential vectors:
\begin{equation}
    \mathbf{n} = \mathbf{t}_{u} \times \mathbf{t}_{v}
\end{equation}
and the normal vector of a point \(\mathbf{x}\) in the screen space can be rendered similar to the color in \cref{eq:color_integration} as:
\begin{equation}\label{eq:normal_integration}
    \hat{N}(\mathbf{x})=\sum_{i \in M}\mathbf{n}_i \alpha_i \prod_{j=1}^{i-1}(1-\alpha_j )
\end{equation}
We omit the low-pass filter \(\hat{\mathcal{G}}\) here and in the following equations for simplicity.

\paragraph{Depth} For the depth, we follow 2DGS to render the expected depth from the depth by intersections of Gaussians and ray by:
\begin{equation}\label{eq:depth_integration}
    D(\mathbf{x})=\sum_{i \in M} d_i \alpha_i \prod_{j=1}^{i-1}(1-\alpha_j ) / \sum_{i \in M}\alpha_i \prod_{j=1}^{i-1}(1-\alpha_j )
\end{equation}

\begin{figure}
    \centering
    \includegraphics[width=1.0\linewidth]{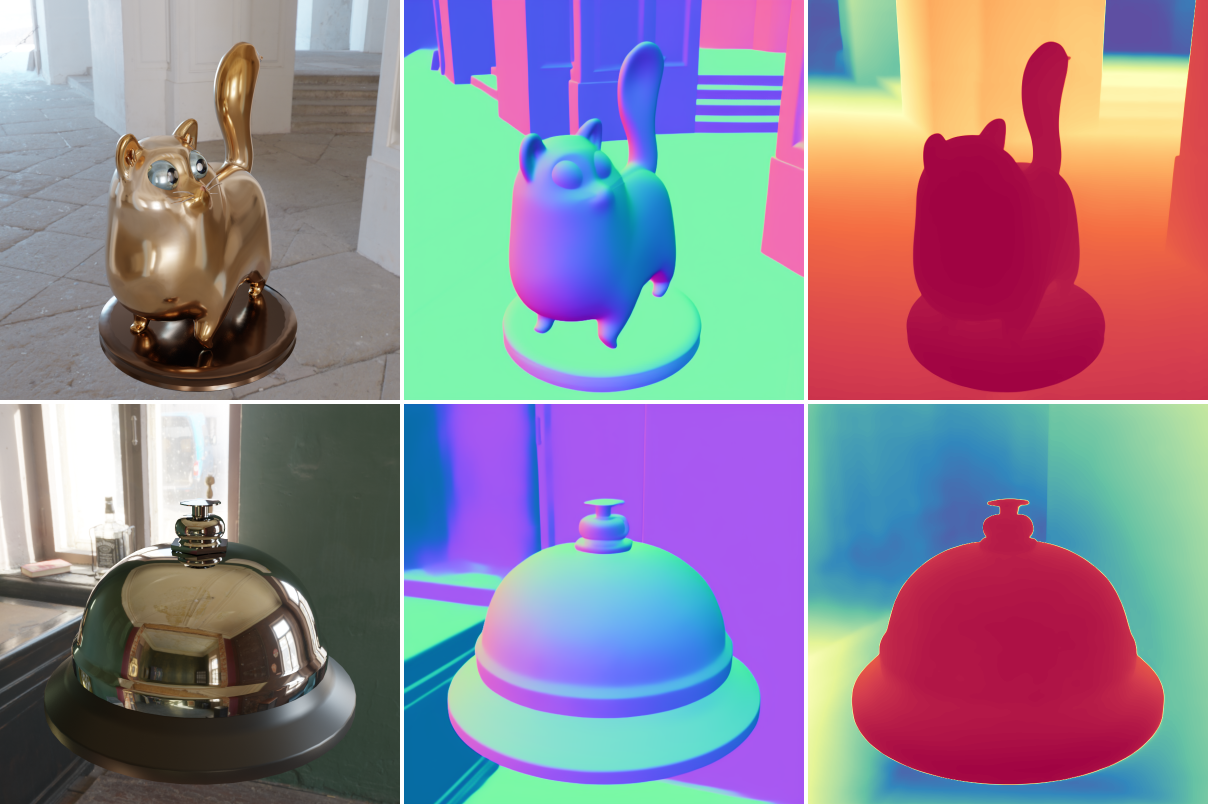}
    \caption{Quality of the normal and depth predictions from the Marigold~\cite{ke2024repurposing,martingarcia2024diffusione2eft} and Depth Pro~\cite{bochkovskii2024depth} foundation models.}
    \label{fig:foundation_model_visual}
\end{figure}

\subsection{Supervision from Foundation Models}\label{sec:foundation_model}
% \TODO{}{Need a clever way to talk about the foundation models.}
As we discussed in the introduction, the monocular depth and normal estimation models infer the geometric information from a single image based on the knowledge of massive training data. Thus, they are insusceptible to the reflective surface problem.
We use the such models~\cite{martingarcia2024diffusione2eft,bochkovskii2024depth} to predict normal \(\tilde{N}\) and depth \(\tilde{D}\) for each image. As shown in \cref{fig:foundation_model_visual}, the foundation models can produce faithful predictions for the object with different materials.

We use the predicted normal \(\tilde{N_i}\) and depth \(\tilde{D_i}\) as pseudo ground-truth for additional supervision. To be specific, we supervise the rendered normal map with \(L_1\) and cosine loss by:
\begin{equation}\label{eq:normal_loss}
    \mathcal{L}_{\text{n}} = \left\| \hat{N}_i - \tilde{N}_i \right\|_1 + \left( 1 - \hat{N}^T \tilde{N} \right)
\end{equation}
As for the depth supervision, we adapt the scale-invariant depth loss~\cite{ranftl2020towards} to supervise the rendered depth map by:
\begin{equation}\label{eq:depth_loss}
    \mathcal{L}_{\text{d}} = \left\| (\omega \hat{D} + b) - \tilde{D} \right\|^2
\end{equation}
where \(\omega\) and \(b\) are the scale and shift used to align the rendered depth \(\hat{D}\) and the predicted depth \(\tilde{D}\). We solve for the \(\omega\) and \(b\) by a least-square optimization according to~\cite{ranftl2020towards}.
% We denote the total geometric loss from foundation models as \(\mathcal{L}_{\text{geo}} = .

\subsection{Shading the Gaussians}
\subsubsection{Recap on Rendering Equation}
In the original Gaussian Splatting, the color of each Gaussian primitive is represented by a view-dependent appearance \(\mathbf{c}\) parameterized with spherical harmonics.
However, the appearance of the reflective surface is not well described by the spherical harmonics. Following previous work \citep{jiang2024gaussianshader,liang2024gs,gao2023relightable}, we introduce physical-based rendering (PBR) pipeline by leveraging the classic rendering equation \cite{kajiya1986rendering} to formulate the radiance at a viewing direction \(\omega_{o}\) of a point \(\mathbf{x}\) as:
\begin{equation}\label{eq:rendering_equation}
    L_{o}(\mathbf{x}, \mathbf{\omega_{o}}, \mathbf{n}) = \int_{\Omega} f_r(\mathbf{x}, \mathbf{\omega_{i}}, \mathbf{\omega_{o}}) L_i(\mathbf{x}, \omega_{i}) (\mathbf{n} \cdot \omega_{i}) \mathrm{d}\omega_{i}
\end{equation}
where \(\Omega\) denotes the upper hemisphere centered at \(\mathbf{x}\), \(\mathbf{n}\) is the normal of the local surface, \(L_i\) is the incoming radiance from the direction \(\omega_{i}\) and \(f_r\) is the bidirectional reflectance distribution function (BRDF).

From \citep{burley2012physically}, the BRDF \(f_r\) can be divided into diffuse and specular components and further decompose according to the Cook-Torrance BRDF model \citep{cook1981reflectance,walter2007microfacet} as follows:
\begin{equation}\label{eq:brdf}
    f_r(\mathbf{\omega_{i}}, \mathbf{\omega_{o}}) = \underbrace{(1 - m) \frac{\mathbf{a}}{\pi}}_{\text{diffuse}} + \underbrace{ \frac{DFG}{4 (\mathbf{\omega_{i}} \cdot \mathbf{n}) (\mathbf{\omega_{o}} \cdot \mathbf{n})}}_{\text{specular}}
\end{equation}
where \(\mathbf{a}, m, \rho\) are the albedo, metallic, and roughness of the surface respectively, \(D\) is the normal distribution function, \(F\) is the Fresnel term, and \(G\) is the geometry term. The computations of \(D, F \text{\ and\ } G\) are related to the surface properties and can be found in \citep{walter2007microfacet}.

According to~\cref{eq:brdf}, the rendering equation~\cref{eq:rendering_equation} can be rewritten as the sum of the diffuse \(L_\mathrm{d}\) and specular \(L_\mathrm{s}\) terms:
\begin{equation}\label{eq:rendering_equation_split}
    \begin{aligned}
         & L_o(\mathbf{x}, \mathbf{\omega_{o}}, \mathbf{n})  = L_{\mathrm{d}} + L_{\mathrm{s}} \text{\ , where}                                                                                                                                                     \\
         & L_{\mathrm{d}} = (1-m) \frac{\mathbf{a}}{\pi} \int_{\Omega} L_i(\mathbf{x}, \mathbf{\omega_i})(\mathbf{\omega_i} \cdot \mathbf{n}) \mathrm{d} \mathbf{\omega_i}                                                                                          \\
         & L_{\mathrm{s}}                       = \int_{\Omega} \frac{D F G}{4 (\mathbf{\omega_{i}} \cdot \mathbf{n}) (\mathbf{\omega_{o}} \cdot \mathbf{n})} L_i(\mathbf{x}, \mathbf{\omega_i})(\mathbf{\omega_i} \cdot \mathbf{n}) \mathrm{d} \mathbf{\omega_i} .
    \end{aligned}
\end{equation}

We follow \citep{munkberg2022extracting} to use a trainable High Dynamic Range (HDR) cube map to represent the environment lighting \(L_{i}(\mathbf{\omega_{i}})\). The diffuse term \(L_\mathrm{d}\) can be pre-computed and stored as a 2D texture map. As for the specular term \(L_\mathrm{s}\), we employ the \textit{split-sum} approximation \citep{karis2013real} to simplify the specular term of the integration.

% From \cref{eq:brdf}, the specular term of the BRDF is view-dependent causing the integration in \cref{eq:rendering_equation} to be intractable. To simplify the calculation, we follow \citep{munkberg2022extracting} to use a learnable cube map to represent environment lighting \(L_{i}(\mathbf{\omega_{i}})\) and employ the split-sum approximation \citep{karis2013real} to simplify the specular term of the integration for computation efficiency.

% To simplify the calculation of the integral in \cref{eq:rendering_equation}, we employ the \textit{split-sum} approximation \citep{karis2013real} as previous works \citep{liu2023nero,mao2023neus-pir,munkberg2022extracting} to approximate the specular term of the integration as. 

% For the specular term of the integration, we employ the \textit{split-sum} approximation \citep{karis2013real}, which has been proved effective in industrial applications and prevous works \citep{liu2023nero,mao2023neus-pir,munkberg2022extracting}, to approximate the integral in \cref{eq:rendering_equation} as:

\begin{figure}[htp]
    \centering
    \includegraphics[width=0.95\linewidth]{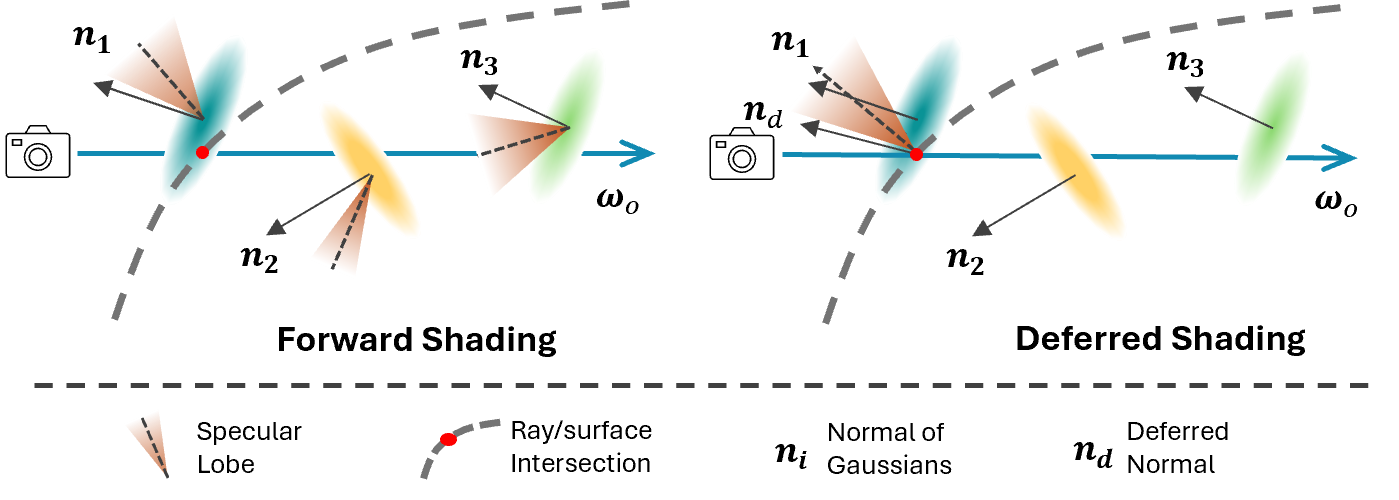}
    \caption{The comparison between forward shading and deferred shading.}
    \label{fig:forward_vs_deferred_shading}
\end{figure}

\begin{figure*}[!t]
    \centering
    \includegraphics[width=\linewidth]{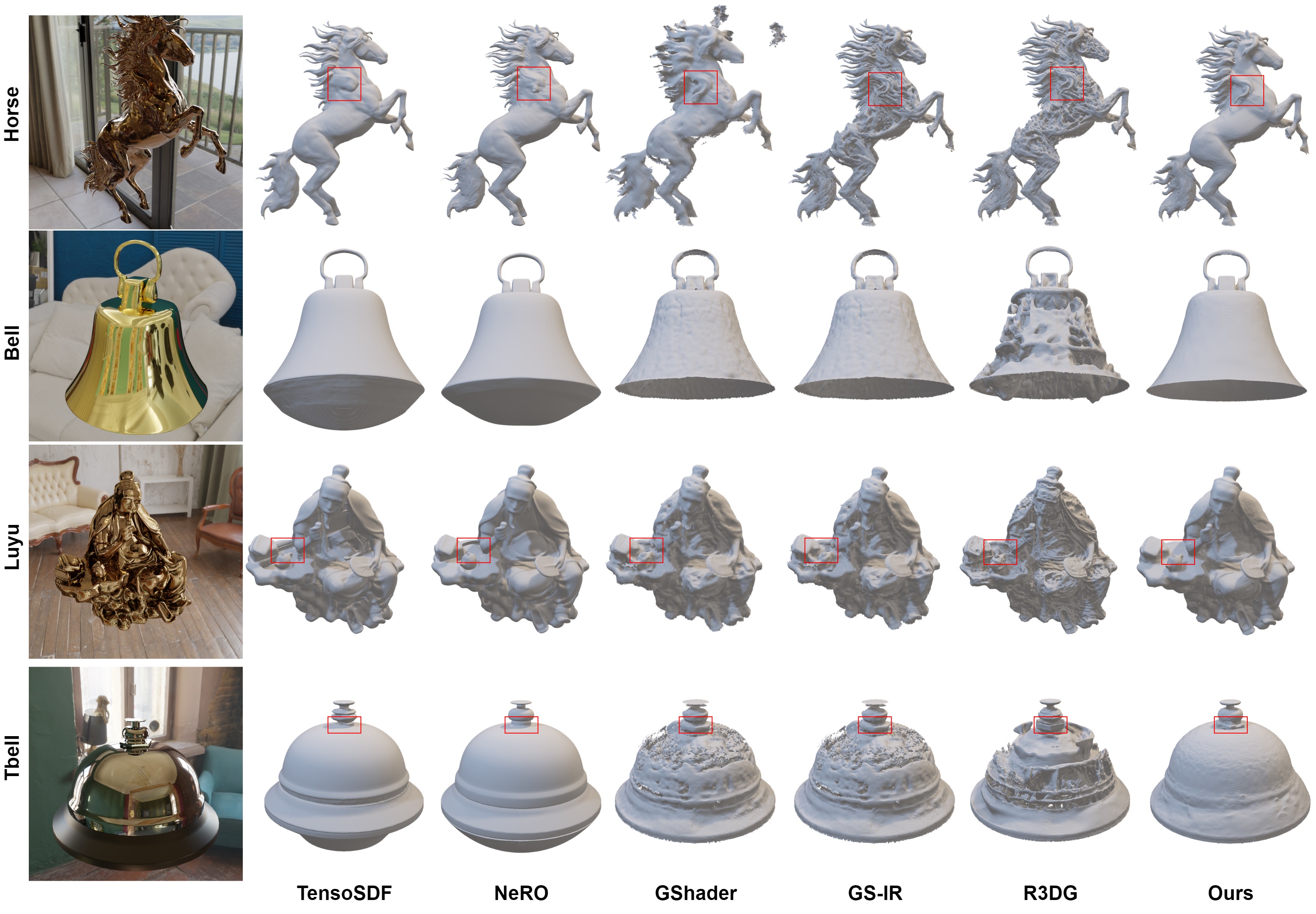}
    \caption{Reconstruction results on the Glossy Bender dataset. Our method achieves significantly better results compared to other Gaussian-based approaches and is comparable to SDF-based methods. We also hold some advantages in finer details than the SDF-based method, as highlighted in the figure.}
    \label{fig:exp_geometry_nerosync}
\end{figure*}

\subsection{Forward Shading vs. Deferred Shading}\label{sec:forward_vs_deferred_shading}
In order to combine 2D Gaussian primitives with the physical-based rendering (PBR) pipeline, we assign each Gaussian primitive an additional set of PBR parameters \((\mathbf{a}, m, \rho)\). A straightforward approach to render the 2D Gaussian primitives by forward shading, in which we calculate the radiance of each Gaussian \(L_o(\mathbf{x}, \mathbf{\omega_{o}}, \mathbf{n})\) of according to \Cref{eq:rendering_equation_split}, and finally get the accumulated color by alpha-blending as \cref{eq:color_integration}.
However, there are several drawbacks: (1) From \cref{eq:rendering_equation_split}, the color of a surface point is determined by the shading point position \(\mathbf{x}\), the normal direction \(\mathbf{n}\), the viewing direction \(\mathbf{\omega_{o}}\) and the PBR parameters. 
In the forward shading, as illustrated in~\cref{fig:forward_vs_deferred_shading}, all of the Gaussians along the camera ray will contribute to the accumulated color. This will cause inaccuracy because each Gaussian can have a different normal direction, position, and it's not essential that they are all aligned with the normal of surface.
(2) The forward shading is computationally expensive because the shading calculation is performed for each Gaussian primitive.

Based on the above analysis, we decide to adopt the deferred shading technique to render the Gaussians. Deferred shading \cite{deering1988triangle} is a rendering technique that decouples the shading and lighting calculations from the geometry rendering. It's firstly introduced to Gaussian Splatting by \citet{ye20243d} to improve the rendering quality in the presence of specular reflections. The rendering process is divided into two stages: the geometry rendering and the shading stage. In the geometry rendering stage, we render the depth, normal and PBR parameters of the scene into the G-buffer. In the shading stage, we only shade the ray once based on the G-buffer information. As illustrated in \cref{fig:forward_vs_deferred_shading}, the deferred shading allows us to shade the ray at the correct shading point and normal direction, producing more accurate shading results.
% The deferred shading technique allows us to render the 2D Gaussian primitives efficiently and also achieve a high-quality environment lighting estimation.

\begin{table*}[htp]
	\centering
	% \setlength{\abovecaptionskip}{5pt}
	% \setlength{\belowcaptionskip}{-5pt}
	% \renewcommand{\arraystretch}{0.85}
	% \resizebox{\textwidth}{!}{
	\begin{tabular}{@{}l|>{\centering\arraybackslash}p{1.3cm}>{\centering\arraybackslash}p{1.3cm}|>{\centering\arraybackslash}p{1.3cm}>{\centering\arraybackslash}p{1.3cm}>{\centering\arraybackslash}p{1.3cm}>{\centering\arraybackslash}p{1.3cm}}
		\toprule[1pt]
		                  & \multicolumn{2}{c|}{SDF-based} & \multicolumn{4}{c}{Gaussian-based}                                                                        \\
		                  & TensoSDF                       & NERO                               & GShader                 & GS-IR  & R3DG   & Ours                     \\
		\midrule
		Angel             & \cellcolor{second}0.0038       & \cellcolor{best}0.0034             & \cellcolor{third}0.0060 & 0.0110 & 0.0090 & 0.0061                   \\
		Bell              & \cellcolor{third}0.0066        & \cellcolor{best}0.0032             & 0.0078                  & 0.1097 & 0.0403 & \cellcolor{second}0.0037 \\
		Cat               & \cellcolor{third}0.0267        & \cellcolor{best}0.0044             & 0.0175                  & 0.0566 & 0.0326 & \cellcolor{second}0.0074 \\
		Horse             & \cellcolor{best}0.0033         & \cellcolor{second}0.0049           & 0.0072                  & 0.0149 & 0.0117 & \cellcolor{third}0.0050  \\
		Luyu              & \cellcolor{third}0.0083        & \cellcolor{best}0.0054             & 0.0101                  & 0.0224 & 0.0151 & \cellcolor{second}0.0076 \\
		Potion            & \cellcolor{second}0.0064       & \cellcolor{best}0.0053             & 0.0382                  & 0.0593 & 0.0380 & \cellcolor{third}0.0088  \\
		Tbell             & \cellcolor{third}0.0212        & \cellcolor{best}0.0035             & 0.0308                  & 0.0989 & 0.0472 & \cellcolor{second}0.0078 \\
		Teapot            & \cellcolor{third}0.0085        & \cellcolor{best}0.0037             & 0.0178                  & 0.0693 & 0.0488 & \cellcolor{second}0.0083 \\
		\midrule
		Average           & \cellcolor{third}0.0106        & \cellcolor{best}0.0042             & 0.0169                  & 0.0553 & 0.0303 & \cellcolor{second}0.0068 \\
		\midrule
		Training Time (h) & 6                              & 12                                 & 0.5                     & 0.5    & 1      & 0.7                      \\
		% Rendering (FPS)   & -                              & -                                  & 50      & 214    & 1.5       & 160    \\
		\bottomrule[1pt]
	\end{tabular}
	\caption{The Chamfer-\(\mathcal{L}_1\) \(\downarrow\) distance of 3D reconstruction results on the Blender Glossy dataset. We color each cell as \colorbox{best}{best}, \colorbox{second}{second}, and \colorbox{third}{third}. Our method achieves top results among GS-based approaches at similar computational cost, and second best for SDF-based approaches at an order of magniture less computational cost.}
	% Masks are generated via segmentation~\citep{kirillov2023sam, ke2024segment,yang2023track} and can be noisy.
	\label{tab:exp_geometry_nerosync}
	% \vspace{-2em}
\end{table*}

\subsection{Training Objective}
We train our model with multiple loss functions. Firstly, we adapt the basic losses \(\mathcal{L}_{\text{GS}}\) from 2DGS~\cite{huang20242dgs} including RGB reconstruction loss and normal consistency loss.
We also introduce the geometric losses \(\mathcal{L}_{\text{n}}\) and \(\mathcal{L}_{\text{d}}\) from the foundation models as we proposed in \cref{sec:foundation_model}.

Finally, we introduce regularization terms on the lighting and PBR parameters to facilitate the training process. We adapt the natural light regularization from~\cite{liu2023nero} by:
\begin{equation}
    \mathcal{L}_{\text{light}} = \left\| \mathbf{L} - \overline{\mathbf{L}} \right\|^2
\end{equation}
where \(\mathbf{L}\) is the predicted environment lighting and \(\overline{\mathbf{L}}\) is the mean of three channels.
We also introduce the regularization on the PBR parameters from~\cite{gao2023relightable} assuming those properties will not change dramatically in the region of smooth color:
\begin{equation}
    \mathcal{L}_{\text{pbr}} = \left\| \nabla \mathbf{X} \right\| \exp(-\left\| \nabla \mathbf{C}_{gt} \right\|)
\end{equation}
where \(\mathbf{X}\) is the rendered map of PBR parameters similar to \cref{eq:color_integration} and \(\mathbf{C}_{gt}\) is the ground-truth color.

In total, the final loss function is formulated as a combination of all the losses:
\begin{equation}
    \mathcal{L} = \mathcal{L}_{\text{GS}} + \lambda_{\text{n}} \mathcal{L}_{\text{n}} + \lambda_{\text{d}} \mathcal{L}_{\text{d}} + \lambda_{\text{light}}\mathcal{L}_{\text{light}} + \lambda_{\text{pbr}}\mathcal{L}_{\text{pbr}}
\end{equation}

% In our method, we adopt the deferred shading technique to render the 2D Gaussian primitives. The rendering process is divided into two stages: the geometry rendering and the shading rendering. In the geometry rendering stage, we render the depth and normal of the scene into the G-buffer. In the shading rendering stage, we render the 2D Gaussian primitives into the scene and calculate the shading result based on the G-buffer information. The deffered shading technique allows us to render the 2D Gaussian primitives efficiently and achieve high-quality rendering results.

\section{Experiments}
\label{sec:experiment}

\begin{table}[h]
    \centering
    \begin{tabularx}{\linewidth}{X|>{\centering\arraybackslash}X >{\centering\arraybackslash}X >{\centering\arraybackslash}X >{\centering\arraybackslash}X}
        \toprule[1pt]
        Methods & Gshader                  & GS-IR  & R3DG                     & Ours                     \\
        \midrule
        Baking  & 0.0056                   & 0.0064 & \cellcolor{second}0.0041 & \cellcolor{best}0.0031   \\
        Ball    & 0.0046                   & 0.0249 & \cellcolor{second}0.0046 & \cellcolor{best}0.0028   \\
        Blocks  & 0.0082                   & 0.0078 & \cellcolor{second}0.0064 & \cellcolor{best}0.0041   \\
        Cactus  & 0.0074                   & 0.0070 & \cellcolor{second}0.0043   & \cellcolor{best}0.0039 \\
        Car     & 0.0062                   & 0.0060 & \cellcolor{second}0.0045   & \cellcolor{best}0.0041 \\
        Chips   & 0.0055                   & 0.0078 & \cellcolor{second}0.0044   & \cellcolor{best}0.0036 \\
        Cup     & 0.0127                   & 0.0138 & \cellcolor{second}0.0108 & \cellcolor{best}0.0092   \\
        Curry   & 0.0053                   & 0.0067 & \cellcolor{second}0.0040   & \cellcolor{best}0.0031 \\
        Gnome   & 0.0067                   & 0.0140 & \cellcolor{second}0.0048   & \cellcolor{best}0.0041 \\
        Grogu   & 0.0151                   & 0.0149 & \cellcolor{second}0.0127 & \cellcolor{best}0.0105   \\
        Pepsi   & \cellcolor{second}0.0065 & 0.0235 & 0.0107                   & \cellcolor{best}0.0035   \\
        Pitcher & \cellcolor{second}0.0062   & 0.0307 & 0.0186                   & \cellcolor{best}0.0072 \\
        Salt    & 0.0078                   & 0.0249 & \cellcolor{second}0.0044   & \cellcolor{best}0.0038 \\
        Teapot  & 0.0056                   & 0.0054 & \cellcolor{second}0.0039 & \cellcolor{best}0.0031   \\
        \midrule
        Average & 0.0074                   & 0.0139 & \cellcolor{second}0.0070 & \cellcolor{best}0.0047   \\
        \bottomrule[1pt]
    \end{tabularx}
    \caption{The reconstruction quality in terms of the Chamfer-\(\mathcal{L}_1\)\(\downarrow\) distance on the Stanford-ORB dataset.}
    \label{tab:exp_geometry_stanfordorb}
\end{table}
\begin{table*}[!htp]
	\centering
	% \resizebox{\linewidth}{!}{
	\begin{tabular}{l|cccc}
		\toprule[1pt]
		Methods         & GShader        & GS-IR          & R3DG                             & Ours                             \\
		\hline
		Angel           & 15.62 / 0.8041 & 17.77 / 0.8162 & \cellcolor{second}19.55 / 0.8279 & \cellcolor{best}19.91 / 0.8510   \\
		Bell            & 15.62 / 0.8316 & 15.69 / 0.8208 & \cellcolor{second}18.71 / 0.8387 & \cellcolor{best}18.95 / 0.8652   \\
		Cat             & 14.51 / 0.8914 & 17.38 / 0.8266 & \cellcolor{second}20.22 / 0.8623 & \cellcolor{best}21.78 / 0.8793   \\
		Horse           & 19.26 / 0.7982 & 20.97 / 0.8980 & \cellcolor{second}21.25 / 0.9025 & \cellcolor{best}22.59 / 0.9190   \\
		Luyu            & 14.41 / 0.7372 & 18.46 / 0.8053 & \cellcolor{best}20.30 / 0.8270   & \cellcolor{second}18.82 / 0.8235 \\
		Potion          & 11.35 / 0.7663 & 15.44 / 0.7601 & \cellcolor{best}19.81 / 0.8192   & \cellcolor{second}17.88 / 0.8299 \\
		Tbell           & 13.37 / 0.8257 & 14.42 / 0.7491 & \cellcolor{second}16.50 / 0.7977 & \cellcolor{best}17.55 / 0.8118   \\
		Teapot          & 15.56 / 0.8108 & 16.78 / 0.8139 & \cellcolor{second}17.22 / 0.8179 & \cellcolor{best}18.96 / 0.8659   \\
		\midrule
		Average         & 14.96 / 0.8108 & 17.11 / 0.8113 & \cellcolor{second}19.19 / 0.8366 & \cellcolor{best}19.56 / 0.8557   \\
		\midrule
		Rendering (FPS) & 50             & 214            & 1.5                              & 160                              \\
		\bottomrule[1pt]
	\end{tabular} % }
	\caption{The relighting quality in terms of PSNR\(\uparrow\) and SSIM\(\downarrow\) on the Blender Glossy dataset, we report the average metrics of three new environment lights. The comparison shows our method achieves the highest rendering quality and competitive framerate.} 
	\label{tab:exp_relighting_nerosync}
\end{table*}

\begin{figure}
    \centering
    \includegraphics[width=\linewidth]{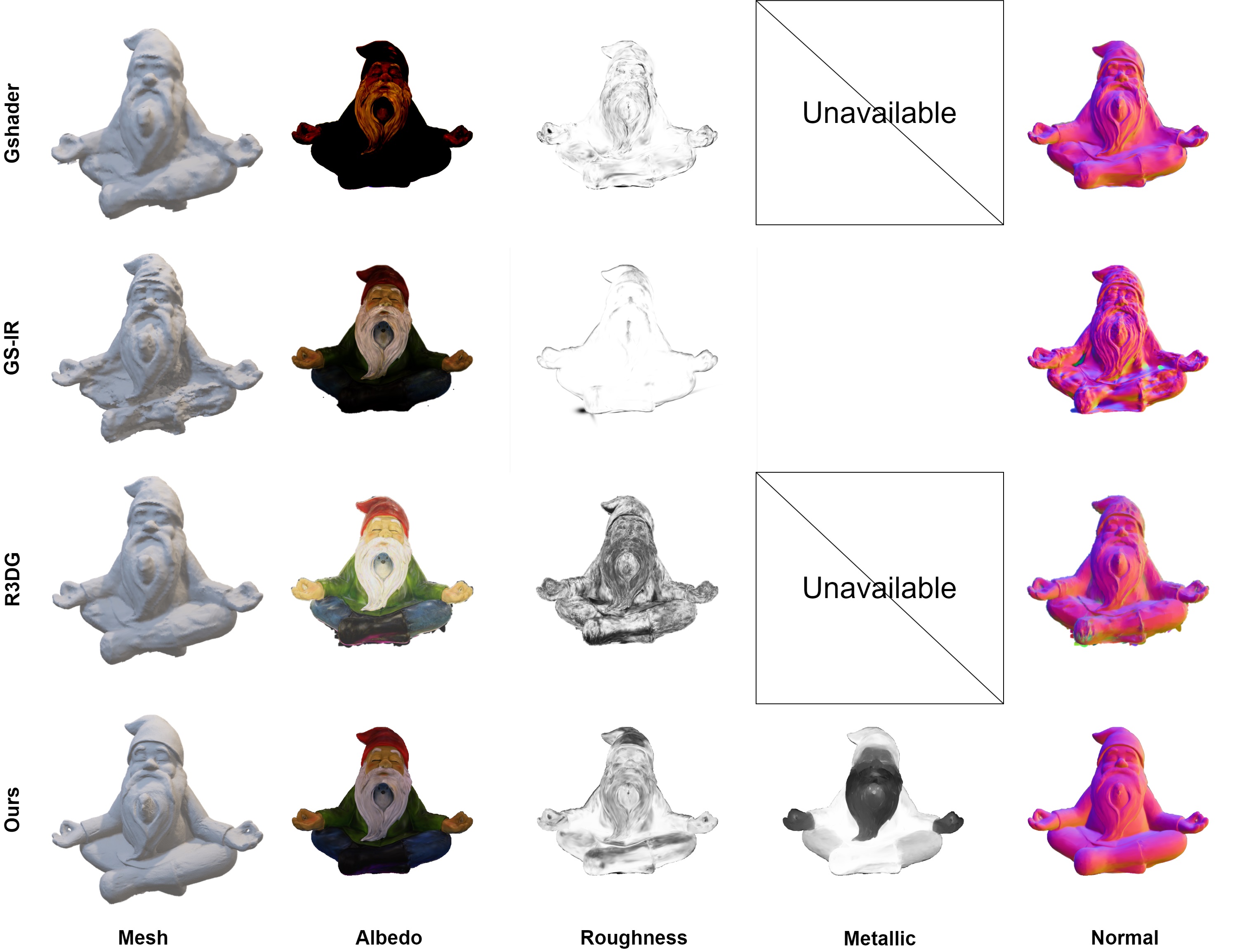}
    \caption{Comparison of the results on the StanfordORB dataset. We show the reconstructed mesh, rendered PBR parameters, and the normal map. Gshader and R3DG do not have Metalic results.}
    \label{fig:exp_geometry_stanfordorb}
\end{figure}

\paragraph{Datasets.}
We conduct experiments on one synthetic dataset Glossy Blender dataset from NeRO~\cite{liu2023nero} and one real-world dataset StanfordORB~\cite{kuang2024stanford}. The Glossy Blender dataset contains 8 objects with highly specular materials, we use it to evaluate the reconstruction and relighting quality. The StanfordORB dataset contains 14 objects with various materials and each object is captured under 3 different lighting conditions. We randomly select one lighting condition from each object. We use the low dynamic range (LDR) images and downsample to \(1024 \times 1024\) for training.

\paragraph{Implementation Details.}
We adopt the Marigold~\cite{ke2024repurposing} fine-tuned by \cite{martingarcia2024diffusione2eft} for normal estimation and Depth Pro~\cite{bochkovskii2024depth} for depth estimation for its superior performance on the object level prediction.
To stabilize the training, we separate the training process into two stages. At the initial stage, we train the original 2DGS model with additional geometric loss \(\mathcal{L}_{\text{n}}\) and \(\mathcal{L}_{\text{d}}\) with \(\lambda_{\text{n}} = 0.5 \) and \(\lambda_{\text{d}} = 0.05 \) for \(30000\) iterations. We follow use the same as 2DGS for other settings. In the second stage, we enable the PBR pipeline and optimize the geometry, PBR parameters and environment lighting together. The weights for the losses are set as \(\lambda_\text{light} = 0.002\),  \(\lambda_\textbf{a} = 0.05\),  \(\lambda_\textbf{m} = 0.05\), and  \(\lambda_\text{r} = 0.01\). The second stage takes another \(10000\) steps to optimize. We implement our method using PyTorch and conduct all the experiments on a single NVIDIA RTX 4090 GPU.

\paragraph{Baselines and metrics.}
For baselines, we select the state-of-the-art methods in the field of reflective object reconstruction and relighting.
We compare our method with the following baselines: \textbf{SDF-based methods}, NeRO~\cite{liu2023nero}: a method based on NeuS~\cite{wang2021neus} and incorporating PBR pipeline, TensoSDF~\cite{li2024tensosdf}: based on NeRO with tensorial representation and roughness-aware training objectives. \textbf{Gaussian-based methods}, GShader~\cite{jiang2024gaussianshader}: a method based on 3DGS and incorporating PBR pipeline, GS-IR~\cite{liang2024gs}: a method improving rendering quality by introducing occlusion and illumination baking, R3DG~\cite{gao2023relightable}: a method further improves relighting quality by point-based ray-tracing. We evaluate the reconstruction quality using the Chamfer distance and the relighting quality using the PSNR and SSIM metrics.

\subsection{Comparison on Reconstruction Quality}
We evaluate the reconstruction quality of our method on the Glossy Bender dataset~\citep{liu2023nero} and the StanfordORB dataset~\citep{kuang2024stanford}. On the Glossy Blender dataset, as shown in \cref{tab:exp_geometry_nerosync}, our method achieves the best reconstruction quality among all the Gaussian-based methods. Compared to the SDF-based method, we achieve comparable results with NeRO~\cite{liu2023nero} and better than TensoSDF~\cite{li2024tensosdf}.
The qualitative comparisons are shown in \cref{fig:exp_geometry_nerosync}. Compared to Gaussian-based methods, our method achieves much better reconstruction quality while others are noisy and bumpy.
The SDF-based methods generally result in smoother surfaces than ours due to the intrinsic smoothness prior of MLP. Our result is not as smooth as theirs, because there is no such constraint from Gaussian. Our method can capture the fine details of the surface such as the horse's mane region and the table bell plunger. We also notice the SDF-based methods tend to produce raised surfaces in the unseen region (bottom of the bells) due to the watertight restriction of SDF and the over-smoothing effect while our method gets clean results in those regions.

As for the results on the StanfordORB dataset shown in \cref{tab:exp_geometry_stanfordorb}, we achieve the best reconstruction quality all over the dataset and our method outperforms other baselines in most cases. From the comparison shown in \cref{fig:exp_geometry_stanfordorb}, our method produce reasonable decomposition of the PBR parameters with less noise.

\subsection{Relighting Comparison}
% We also evaluate the relighting quality of our method on the Glossy Bender dataset~\cite{liu2023nero} shown in \cref{tab:exp_relighting_nerosync}. Our method achieves the best relighting quality among all the Gaussian-based methods. The qualitative comparisons shown in \cref{fig:exp_relighting_nerosync} also show that our method can correctly reconstruct the geometry as well as PBR parameters to enable realistic relighting effect. 

\begin{figure*}[h]
    \centering
    \includegraphics[width=0.9\linewidth]{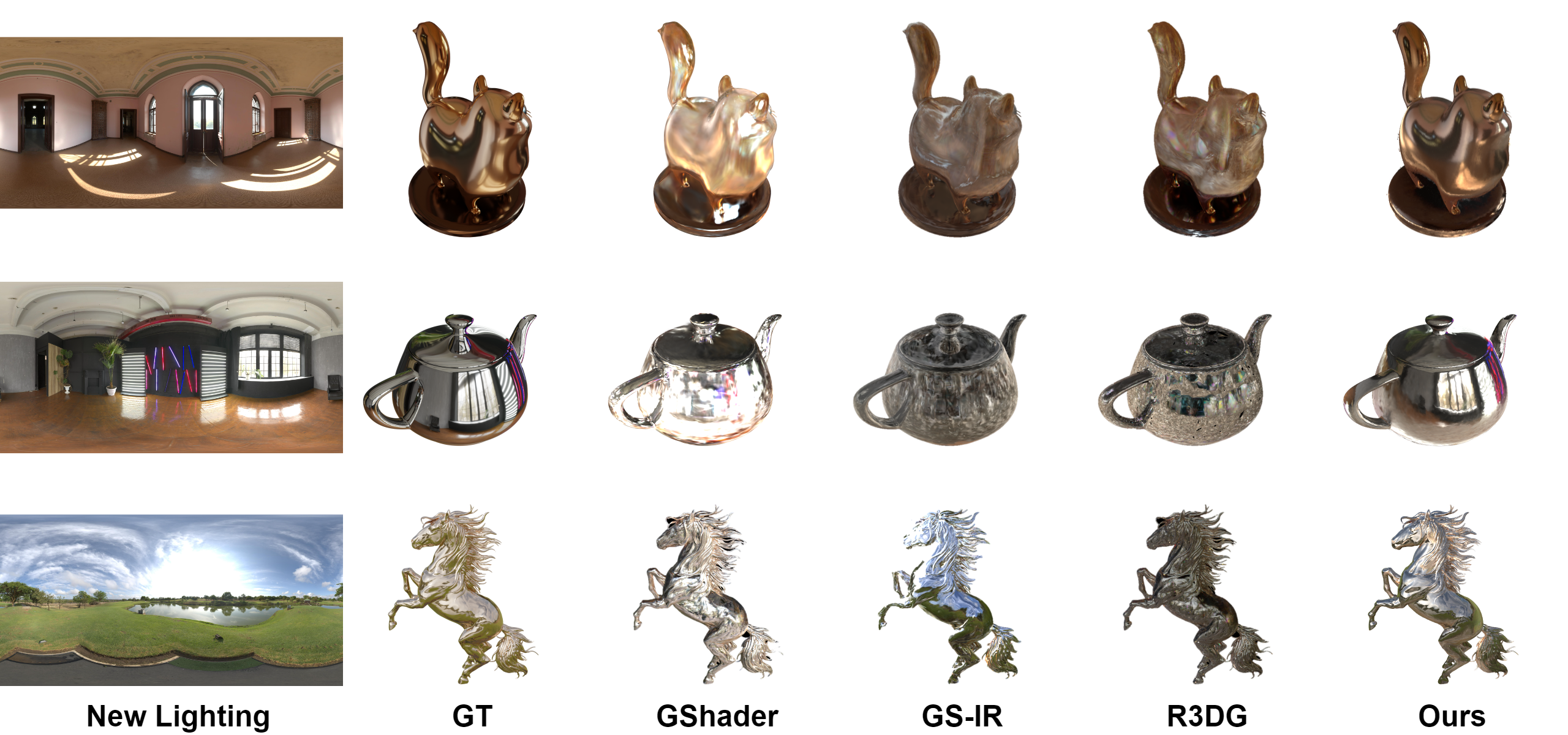}
    \caption{Relighting results on the Glossy Bender dataset.}
    \label{fig:exp_relighting_nerosync}
\end{figure*}

We further evaluate the relighting quality of our method on the Glossy Bender dataset~\cite{liu2023nero}, as presented in \cref{tab:exp_relighting_nerosync}. Our method achieves the highest relighting quality among all Gaussian-based methods. Our approach achieves the highest relighting quality among all Gaussian-based methods. The qualitative comparisons in~\cref{fig:exp_relighting_nerosync} further illustrate that, leveraging the PBR parameters and geometry from our method, we achieve compelling relighting effects, whereas other methods fail to produce reasonable results.

\subsection{Ablation Study}
We validate the effectiveness of each component in our method by conducting ablation studies on the Glossy Bender dataset~\citep{liu2023nero} shown in \cref{tab:exp_ablation}. Starting from the baseline method 2DGS~\cite{huang20242dgs}, we add the geometric supervision from foundation models. The reconstruction quality is improved significantly with the additional supervision. However the rendering quality drops due to the lack of flexibility. We then add the PBR to model the reflective effect, achieving improvement in terms of both reconstruction and rendering quality.

\begin{figure}
    \centering
    \includegraphics[width=\linewidth]{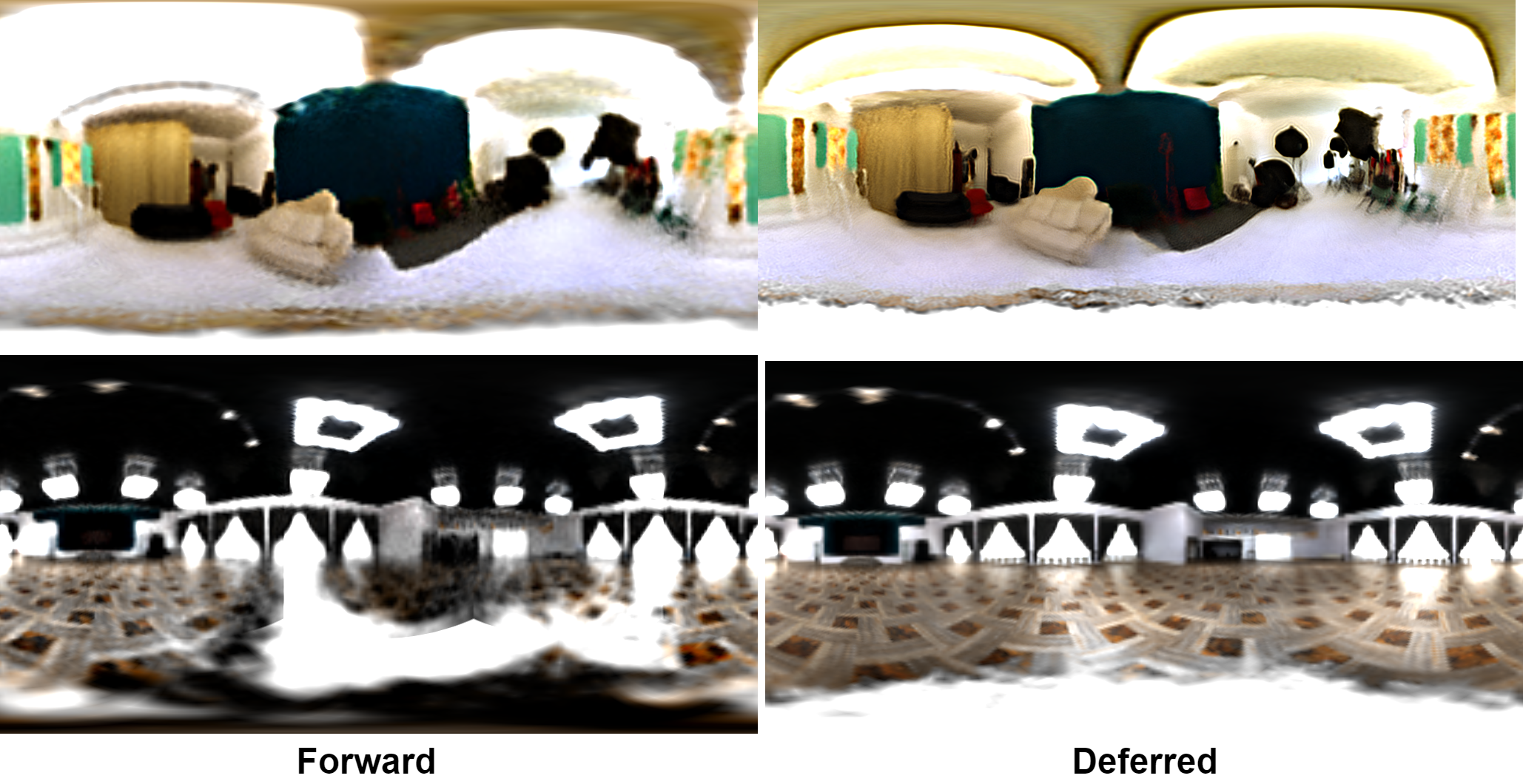}
    \caption{Comparison of environment lighting estimation quality between the deferred shading and forward shading.}
    \label{fig:exp_deferred_vs_forward}
\end{figure}

Finally, we add the deferred shading forming the full model. As we analyze in \cref{sec:forward_vs_deferred_shading}, the deferred shading provides more accurate shading components and thus improves the environment lighting estimation. We visualize the estimated environment lighting in \cref{fig:exp_deferred_vs_forward}. The deferred shading provides a more accurate estimation of the environmental lighting, which leads to better rendering quality.

\begin{table}
	\centering
	\begin{tabular}{cccc|c|c}
		\toprule[1pt]
		2DGS       & Geo        & PBR        & DS         & C-\(\mathcal{L}_1\) \(\downarrow\) & PSNR \(\uparrow\)       \\
		\midrule
		\checkmark &            &            &            & 0.0481                             & \cellcolor{second}26.23 \\
		\midrule
		\checkmark & \checkmark &            &            & \cellcolor{third}0.0084            & 25.52                   \\
		\midrule
		\checkmark & \checkmark & \checkmark &            & \cellcolor{second}0.0074           & \cellcolor{third}25.86  \\
		\midrule
		\checkmark & \checkmark & \checkmark & \checkmark & \cellcolor{best}0.0068             & \cellcolor{best}26.76   \\
		% \checkmark & \checkmark &         &         &         & 0.0101                             & 25.6214           \\
		\bottomrule[1pt]
	\end{tabular}
	\caption{Ablation study results on the Glossy Bender dataset. The result shows that geometric supervision from the foundation model has the most significant impact on the performance of reconstruction and the deferred shading technique helps with rendering quality.}
	\label{tab:exp_ablation}
\end{table}

\section{Conclusion}
\label{sec:conclusion}

% In this paper, we propose GS-2DGS, a Gaussian-based reconstruction method specific to the reflective objects. By introducing physical-based rendering pipeline, our model is able to produce the interaction between surface and environment lighting, thus can mimic the reflective effect of those objects. By leveraging additional geometry supervision from foundation models and introducing the deferred shading technique, our method is able to produce satisfactory geometry reconstruction and relighting effects. However, compared to the state-of-the-art SDF-based methods, we still have a slight gap. We believe this is because the Gaussian-based method inherently lacks geometric constraints. While some concurrent works using SDF to provide additional geometry supervision, those methods usually slow down the running speed. It will be promising future work on how to reconstruct fast and accurate. 

In this paper, we introduce GS-2DGS, a Gaussian-based reconstruction method designed specifically for reflective objects. By incorporating a physics-based rendering pipeline, our model captures the interaction between surfaces and environmental lighting, effectively mimicking the reflective properties of such objects. Leveraging additional geometric supervision from foundation models and incorporating deferred shading, our method achieves high-quality geometry reconstruction and realistic relighting effects at low computational cost.

Despite these advancements and significantly lower computational costs, our method still shows a slight performance gap compared to state-of-the-art SDF-based approaches. We attribute this to the inherent lack of geometric constraints in Gaussian-based methods. While some concurrent works have explored using SDFs to provide additional geometric supervision, these methods typically suffer from high computational costs. Future works could focus on achieving both speed and accuracy in reconstruction, presenting an exciting avenue for further research.

{
    \small
    \bibliographystyle{ieeenat_fullname}
    \bibliography{main}
}

\clearpage
\setcounter{page}{1}
\setcounter{figure}{0}
\setcounter{section}{0}
\renewcommand{\thefigure}{S\arabic{figure}}
\renewcommand{\thetable}{S\arabic{table}}
\maketitlesupplementary

\begin{figure}[htp]
    \centering
    \includegraphics[width=\linewidth]{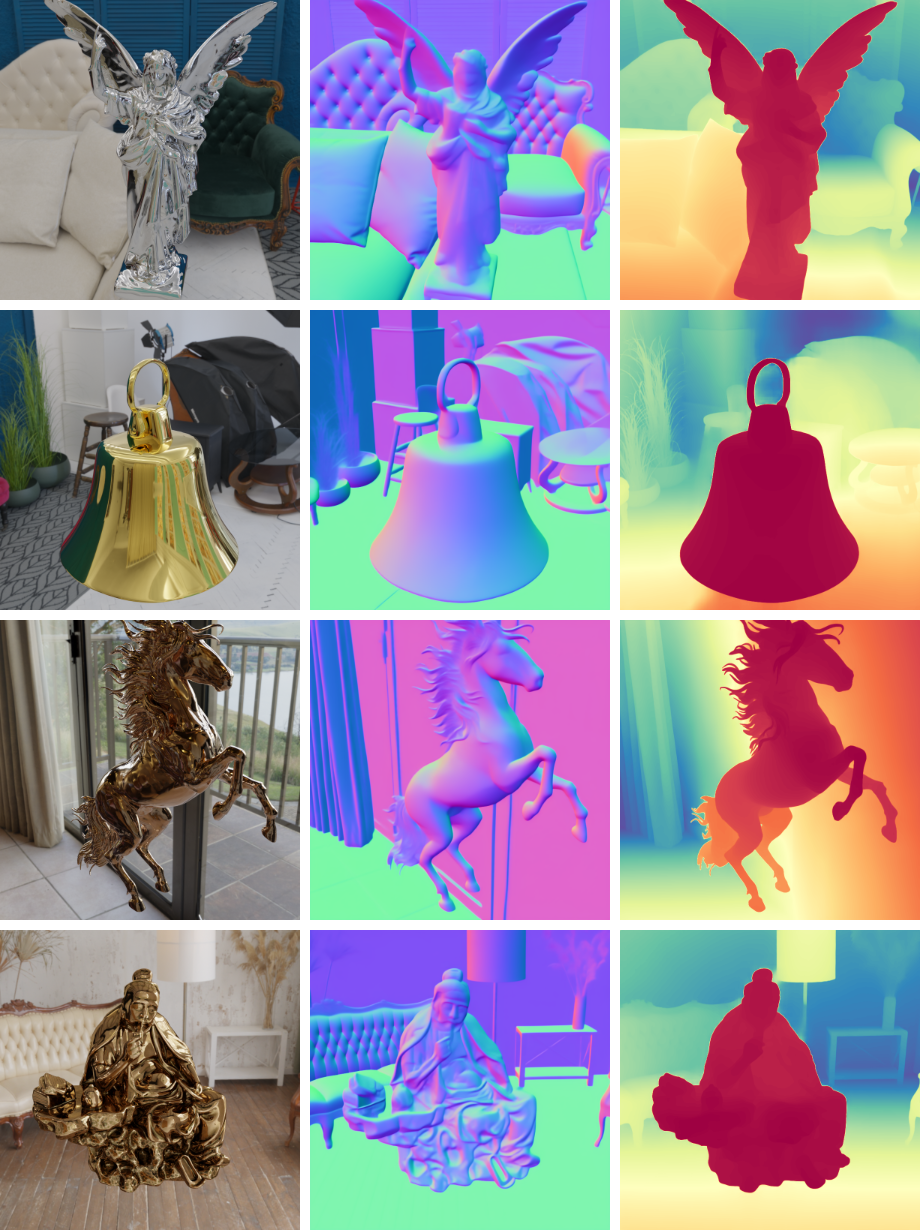}
    \caption{Normal and depth estimation results on the Glossy Blender dataset.}
    \label{fig:supp_foundation_model_nerosync}
\end{figure}

\section{Additional Implementation Details}

\subsection{Representation}
We implement GS-2DGS mainly based on the original 2DGS. The 2DGS uses \(\left\{\mathbf{x}, \mathbf{s}, \mathbf{t}\right\}\) and  \(\left\{\alpha, \mathbf{c}\right\}\) to represent its geometric and volumetric appearance properties respectively, where \(\mathbf{x}\) is the position of the Gaussian, \(\mathbf{s}\) and \(\mathbf{t}\) are the scale and rotation of the axis, \(\alpha\) is the opacity and \(\mathbf{c}\) is the SH coefficients. Apart from those properties, we add PBR parameters \(\left\{\mathbf{a}, \mathbf{m}, \text{and} \mathbf{r}\right\}\) to represent the albedo, metalness, and roughness of the Gaussian.

\subsection{Usage of Foundation Models}
\paragraph{Normal Estimation.}
We adopt the pre-trained model of Marigold~\cite{ke2024repurposing} fine-tuned by \cite{martingarcia2024diffusione2eft} for normal estimation for its robust performance on the reflective objects. We follow the original code \footnote{\url{https://github.com/VisualComputingInstitute/diffusion-e2e-ft}} and default pipeline to use the RGB images as input and predict the normal map.
\paragraph{Depth Estimation.}
For depth estimation, we use the Depth Pro~\cite{bochkovskii2024depth} model for its superior performance on the object level prediction. We use the original implementation \footnote{\url{https://github.com/apple/ml-depth-pro}} and follow the original pipeline to predict the depth map. Specially, for the StanfordORB dataset, we use the masked RGB images without the background for better performance.

\begin{figure}[htp]
    \centering
    \includegraphics[width=\linewidth]{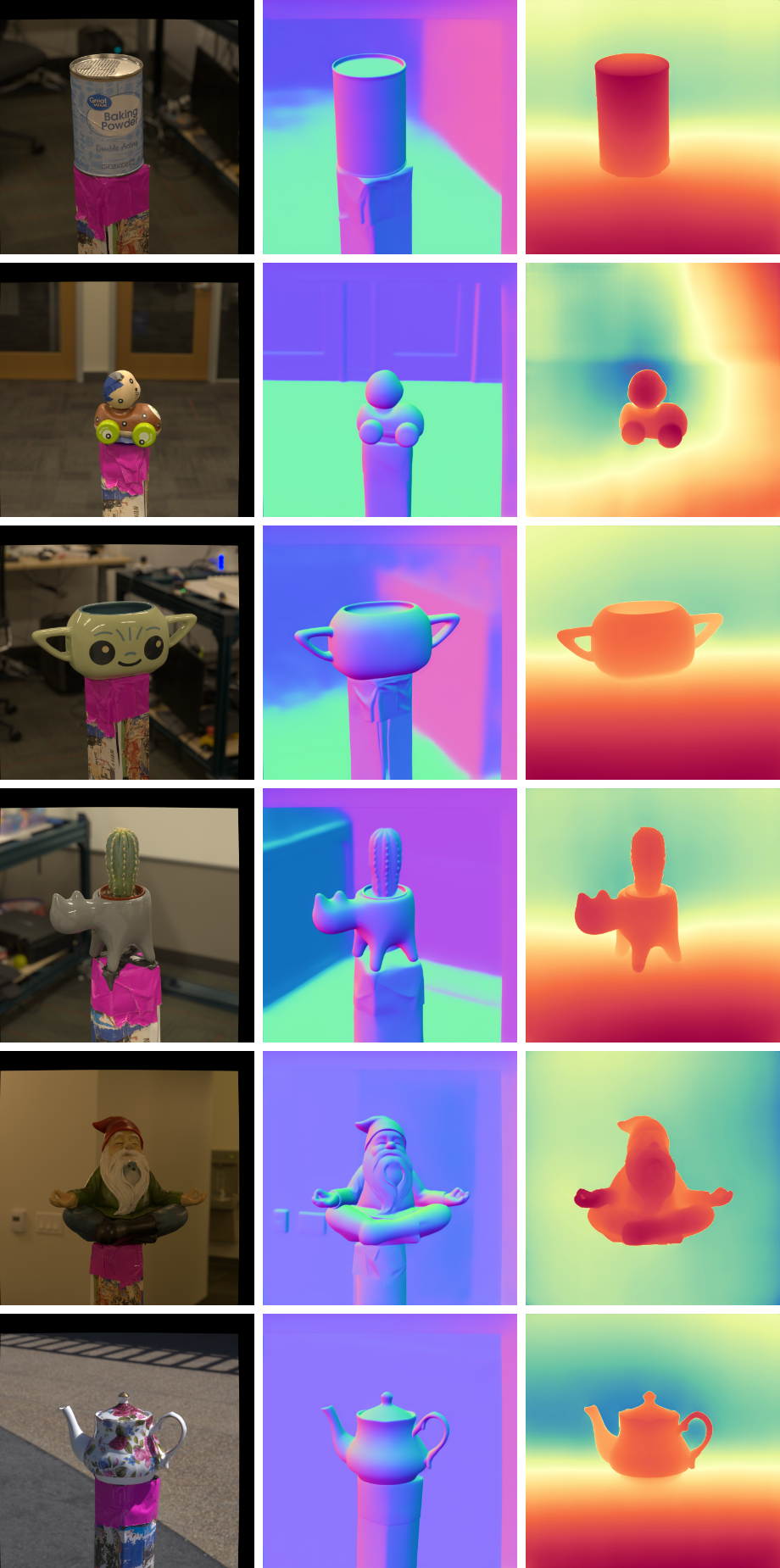}
    \caption{Normal and depth estimation results on the StanfordORB dataset.}
    \label{fig:supp_foundation_model_stanfordorb}
\end{figure}

We show some examples of the normal and depth estimation results in \cref{fig:supp_foundation_model_nerosync} and \cref{fig:supp_foundation_model_stanfordorb}.

% \subsection{Glossy Blender dataset}

\subsection{StanfordORB dataset}

\begin{table}[ht]
    \centering
    \begin{tabular}{l | c | c}
        \toprule[1pt]
        Object  & Training Scene & Evaluation Scenes  \\
        \midrule
        Baking  & scene001       & scene002, scene003 \\
        Ball    & scene003       & scene002, scene004 \\
        Blocks  & scene005       & scene002, scene006 \\
        Cactus  & scene001       & scene005, scene007 \\
        Car     & scene004       & scene002, scene006 \\
        Chips   & scene003       & scene002, scene004 \\
        Cup     & scene006       & scene003, scene007 \\
        Curry   & scene001       & scene005, scene007 \\
        Gnome   & scene003       & scene005, scene007 \\
        Grogu   & scene001       & scene002, scene003 \\
        Pepsi   & scene003       & scene002, scene004 \\
        Pitcher & scene007       & scene001, scene005 \\
        Salt    & scene007       & scene004, scene005 \\
        Teapot  & scene002       & scene001, scene006 \\
        \bottomrule[1pt]
    \end{tabular}
    \caption{Training and evaluation scenes for each object in the StanfordORB dataset.}
    \label{tab:supp_stanfordorb_scene}
\end{table}

The StanfordORB dataset contains 14 objects each in three scenes with different lighting conditions. Both low dynamic range (LDR) and high dynamic range (HDR) images are offered.
For each object, we randomly picked one scene for training and the other two for evaluation as illustrated in \cref{tab:supp_stanfordorb_scene}. The LDR images are downsampled to \(1024 \times 1024\) for training.
% We show the reconstruction results on the StanfordORB dataset in \cref{tab:supp_stanfordorb} and \cref{fig:supp_stanfordorb}.

\begin{figure*}[htp]
    \centering
    \includegraphics[width=\linewidth]{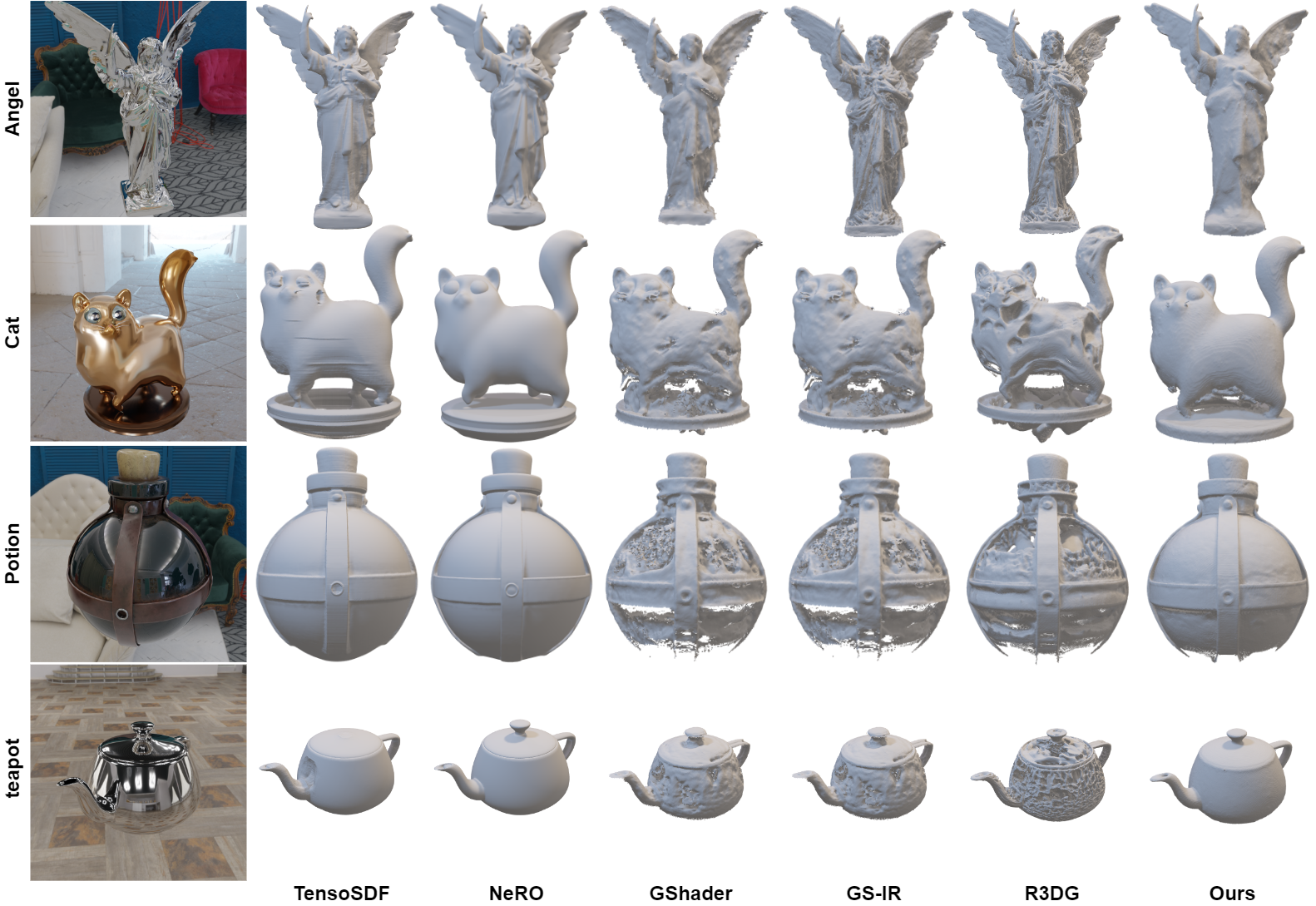}
    \caption{Additional reconstruction results on the Glossy Blender dataset.}
    \label{fig:supp_exp_geometry_nerosync}
\end{figure*}

\begin{figure*}
    \centering
    \includegraphics[width=0.8\linewidth]{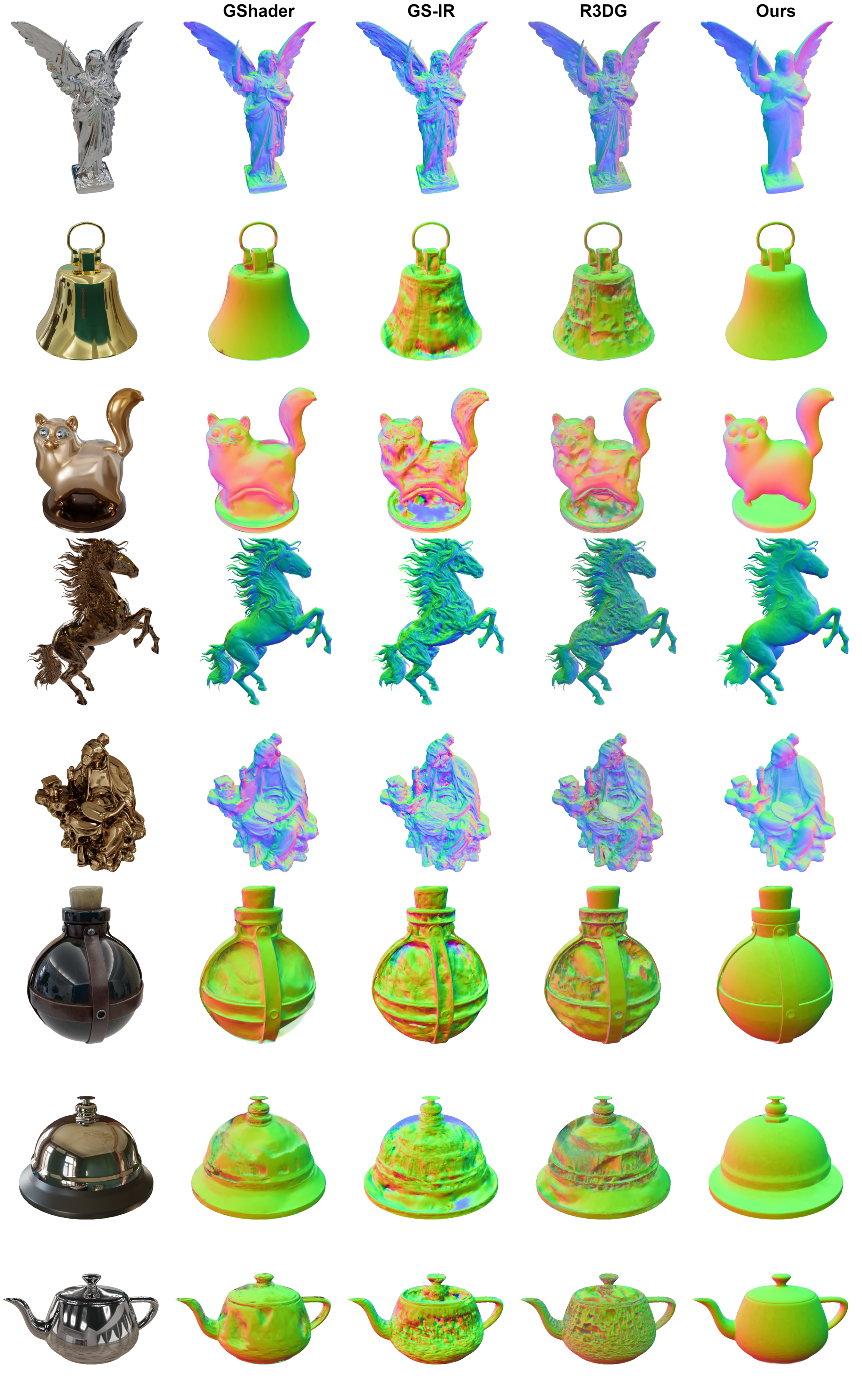}
    \caption{Rendered normal maps results on the Glossy Blender dataset.}
    \label{fig:supp_exp_normal_nerosync}
\end{figure*}

\begin{figure*}[htp]
    \centering
    \includegraphics[width=0.85\linewidth]{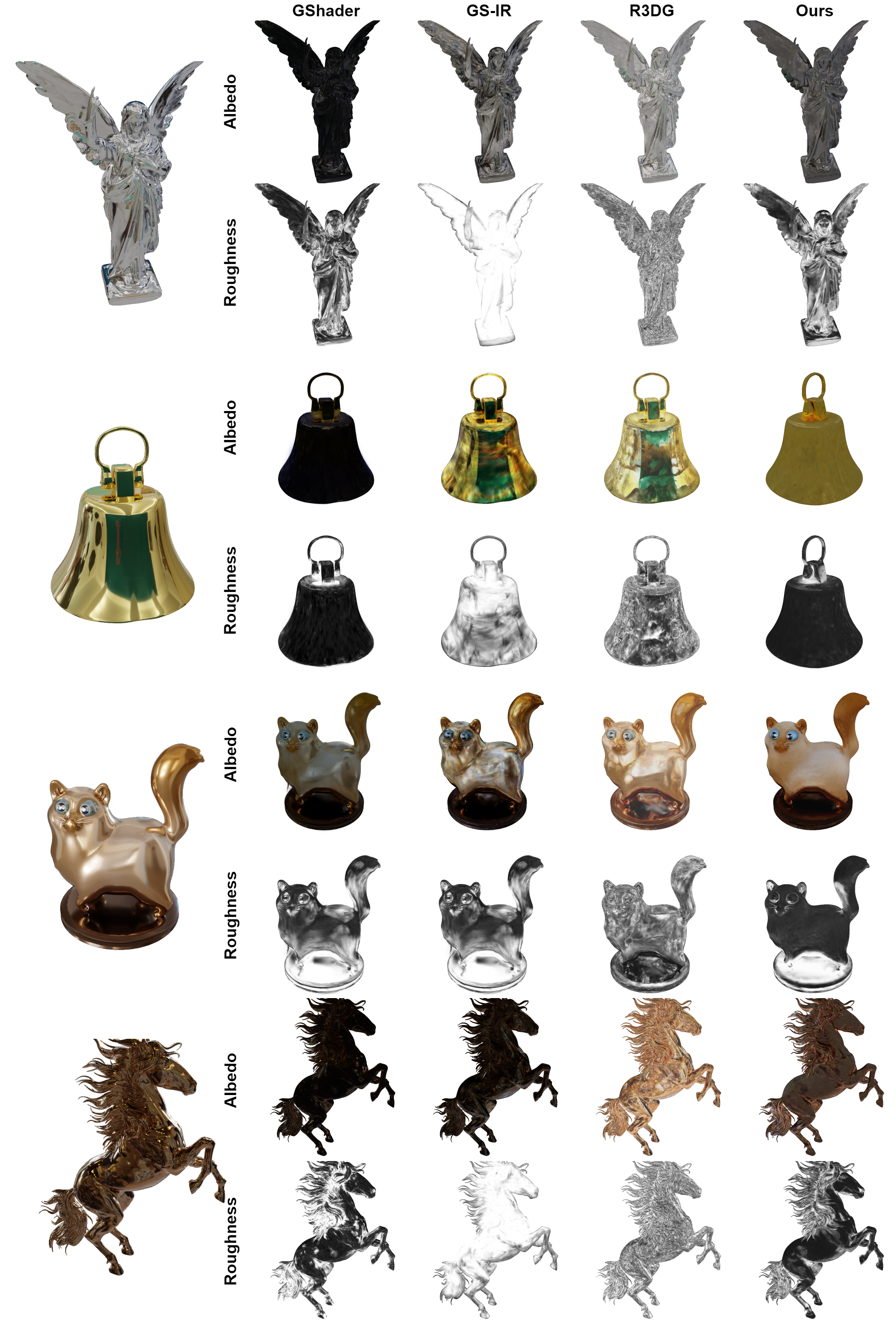}
    \caption{PBR material parameters decomposition on the Glossy Blender dataset.}
    \label{fig:supp_exp_decomposition_nerosync_a}
\end{figure*}

\begin{figure*}
    \centering
    \includegraphics[width=0.85\linewidth]{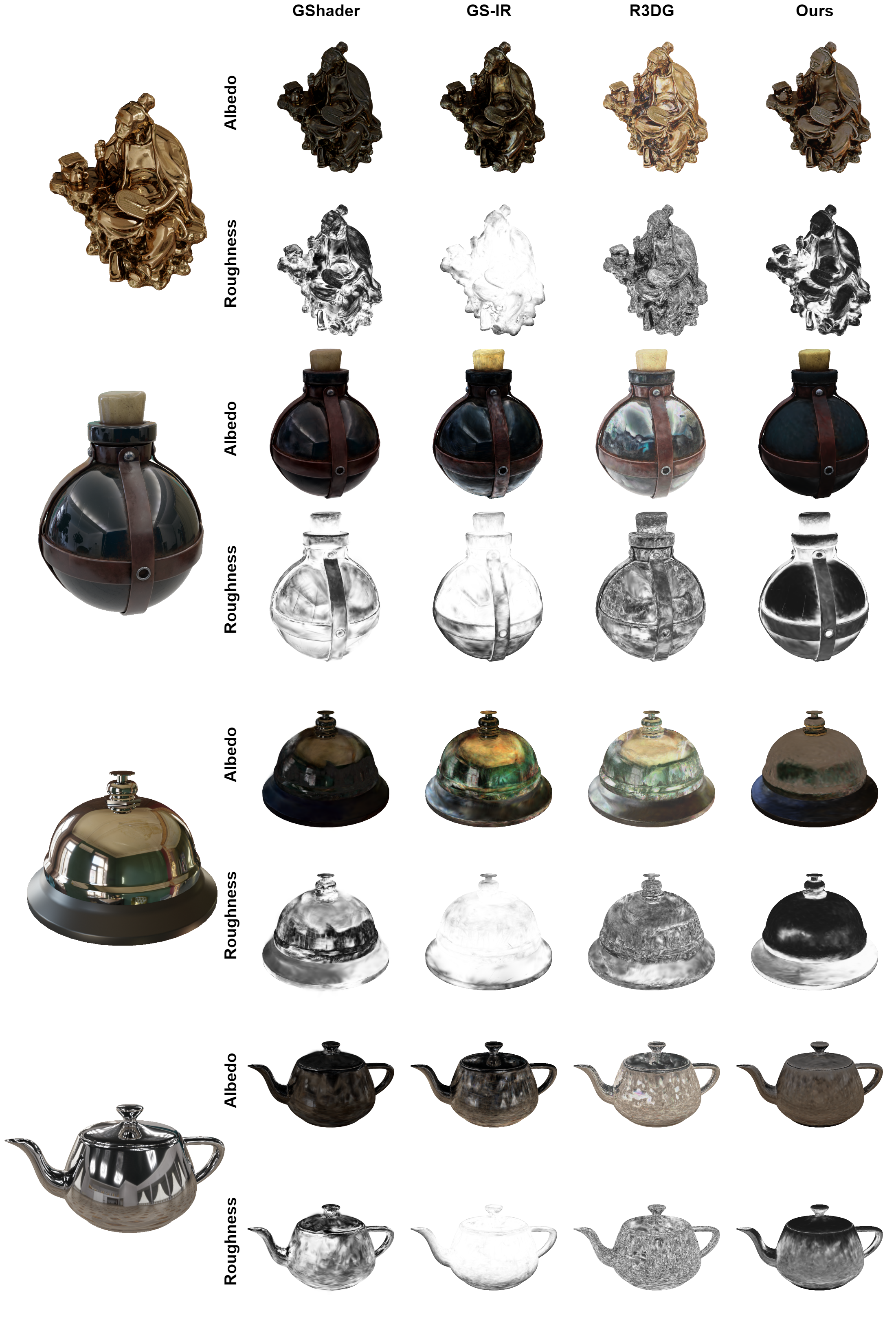}
    \caption{PBR material parameters decomposition on the Glossy Blender dataset.}
    \label{fig:supp_exp_decomposition_nerosync_b}
\end{figure*}

\begin{figure*}
    \centering
    \includegraphics[width=\linewidth]{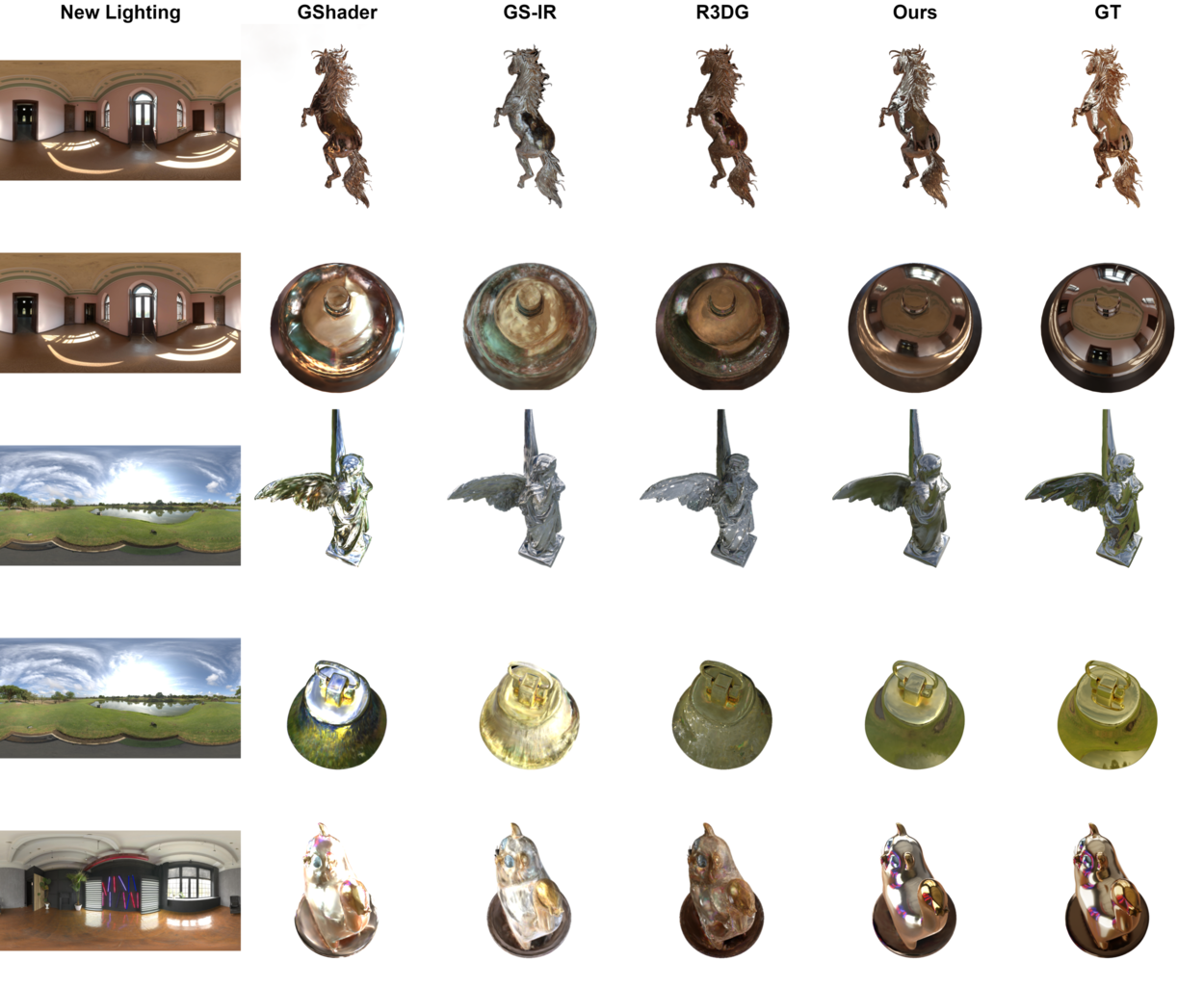}
    \caption{Relighting results on the Glossy Blender dataset.}
    \label{fig:supp_exp_relighting_nerosync}
\end{figure*}

\section{Additional Experiments}

\subsection{Results on Glossy Blender dataset}
\paragraph{Reconstruction} \cref{fig:supp_exp_geometry_nerosync} shows the additional reconstruction results on the Glossy Blender dataset. Our method achieves the best reconstruction quality among all the Gaussian-based methods. Compared to the SDF-based method, we achieve comparable results with NeRO~\cite{liu2023nero} and better than TensoSDF~\cite{li2024tensosdf}. We also show the normal map results in \cref{fig:supp_exp_normal_nerosync}.

\paragraph{Material Decomposition and Relighting.} We show the PBR related material parameters decomposition in \cref{fig:supp_exp_decomposition_nerosync_a} and \cref{fig:supp_exp_decomposition_nerosync_b}. We render the corresponding parameter map by alpha blending according to Eq.(3) of the main paper. % \cref{eq:color_integration}. 
Compared to other methods, our method can get a more reasonable decomposition of the PBR parameters with less noise. We also show the relighting results of different objects under several environment lights in \cref{fig:supp_exp_relighting_nerosync}.
\subsection{Results on StanfordORB dataset}
\begin{figure*}
    \centering
    \includegraphics[width=0.85\linewidth]{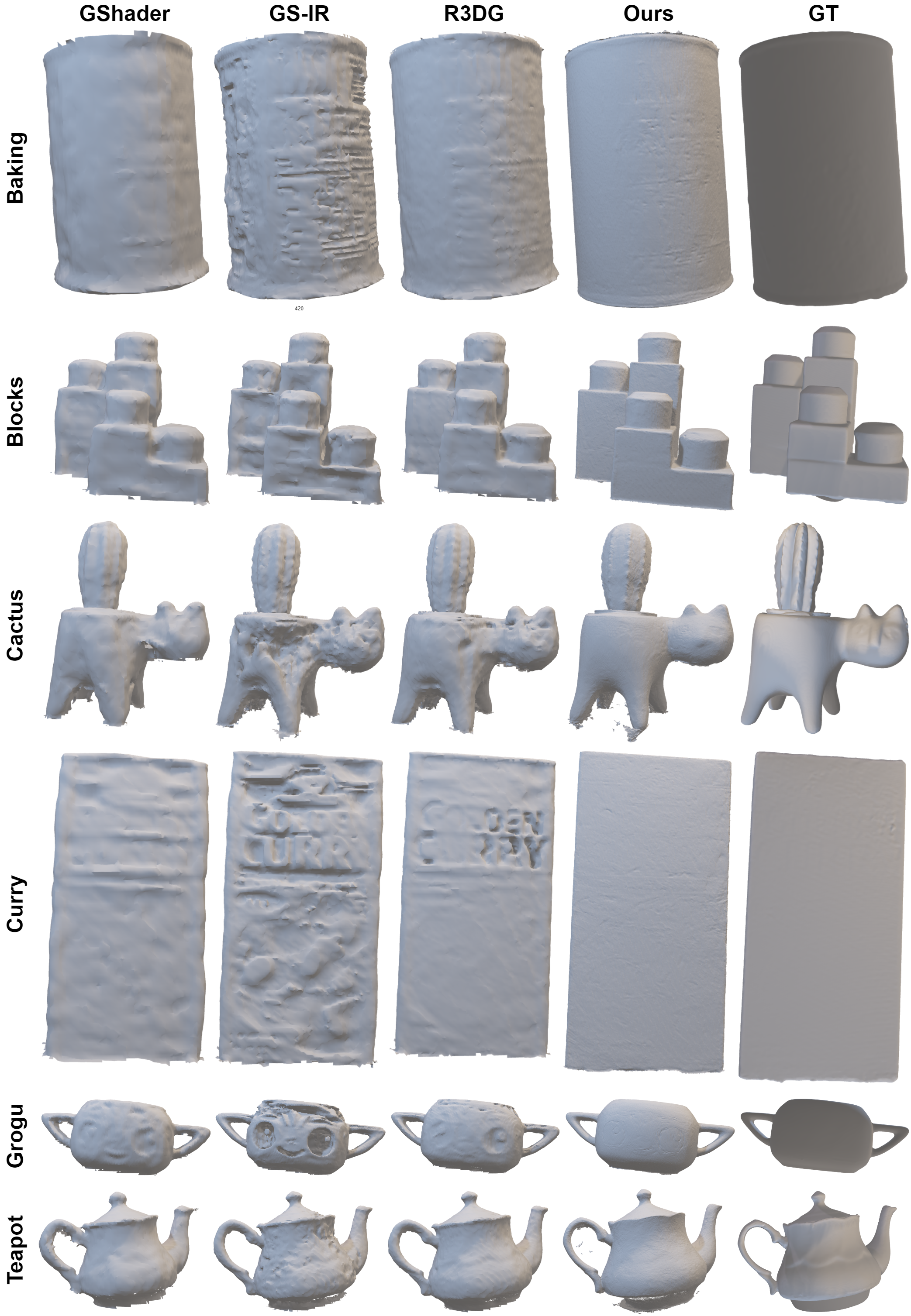}
    \caption{Additional reconstruction results on the StanfordORB dataset.}
    \label{fig:supp_exp_geometry_stanfordorb}
\end{figure*}

Here, we provide more details of the experiment results on the StanfordORB dataset.
\paragraph{Reconstruction} In \cref{fig:supp_exp_geometry_stanfordorb}, we show the comparison of reconstructed geometry among different methods. Our method achieves faithful reconstruction results with less noise and smooth surfaces.
% We show the reconstruction results on the StanfordORB dataset in \cref{tab:supp_stanfordorb} and \cref{fig:supp_stanfordorb}. Our method achieves the best reconstruction quality among all the Gaussian-based methods. Compared to the SDF-based method, we achieve comparable results with NeRO~\cite{liu2023nero} and better than TensoSDF~\cite{li2024tensosdf}.

\begin{figure*}
    \centering
    \includegraphics[width=0.9\linewidth]{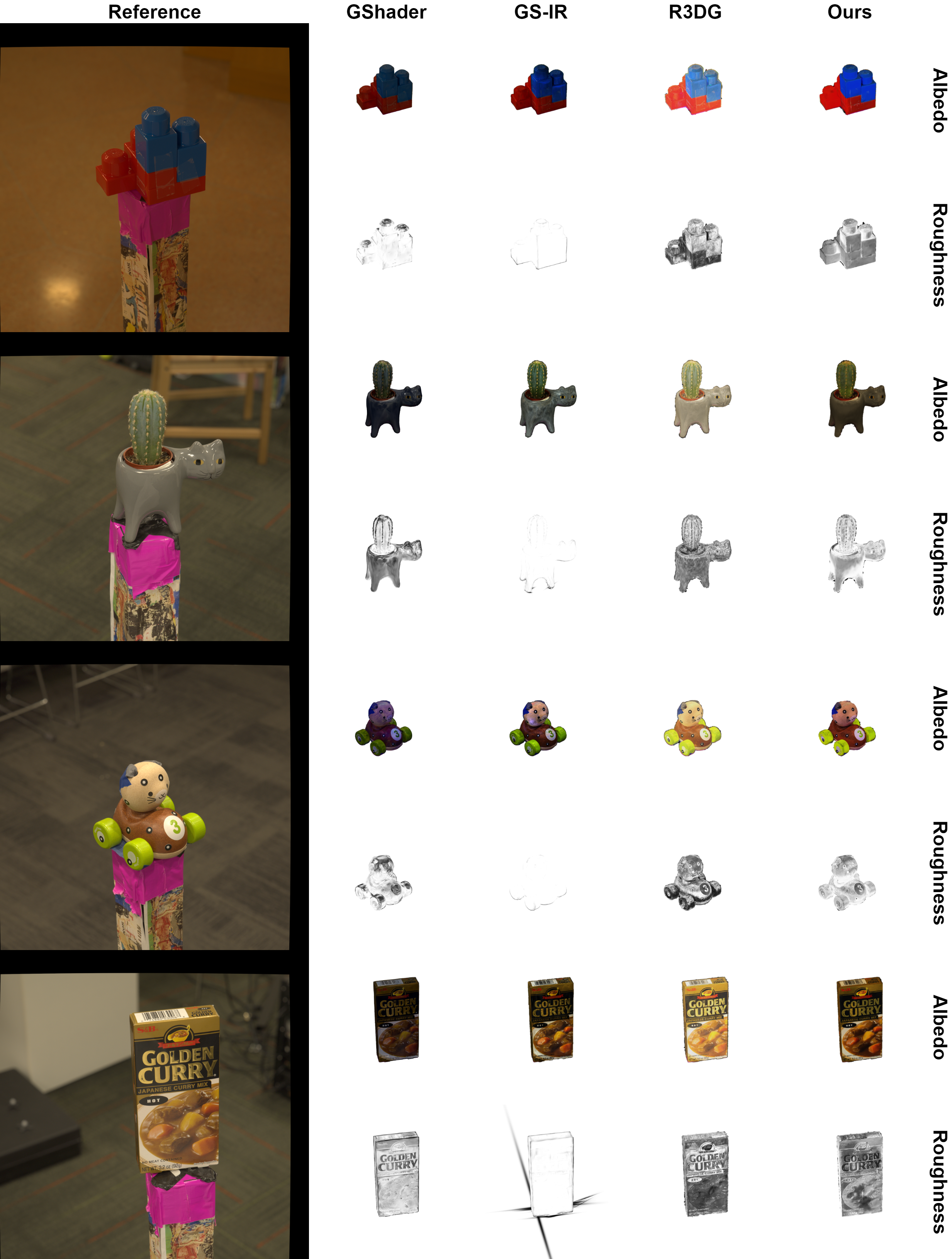}
    \caption{PBR material decomposition on the StanfordORB dataset.}
    \label{fig:supp_exp_decomposition_stanfordorb}
\end{figure*}

\begin{figure*}
    \centering
    \includegraphics[width=\linewidth]{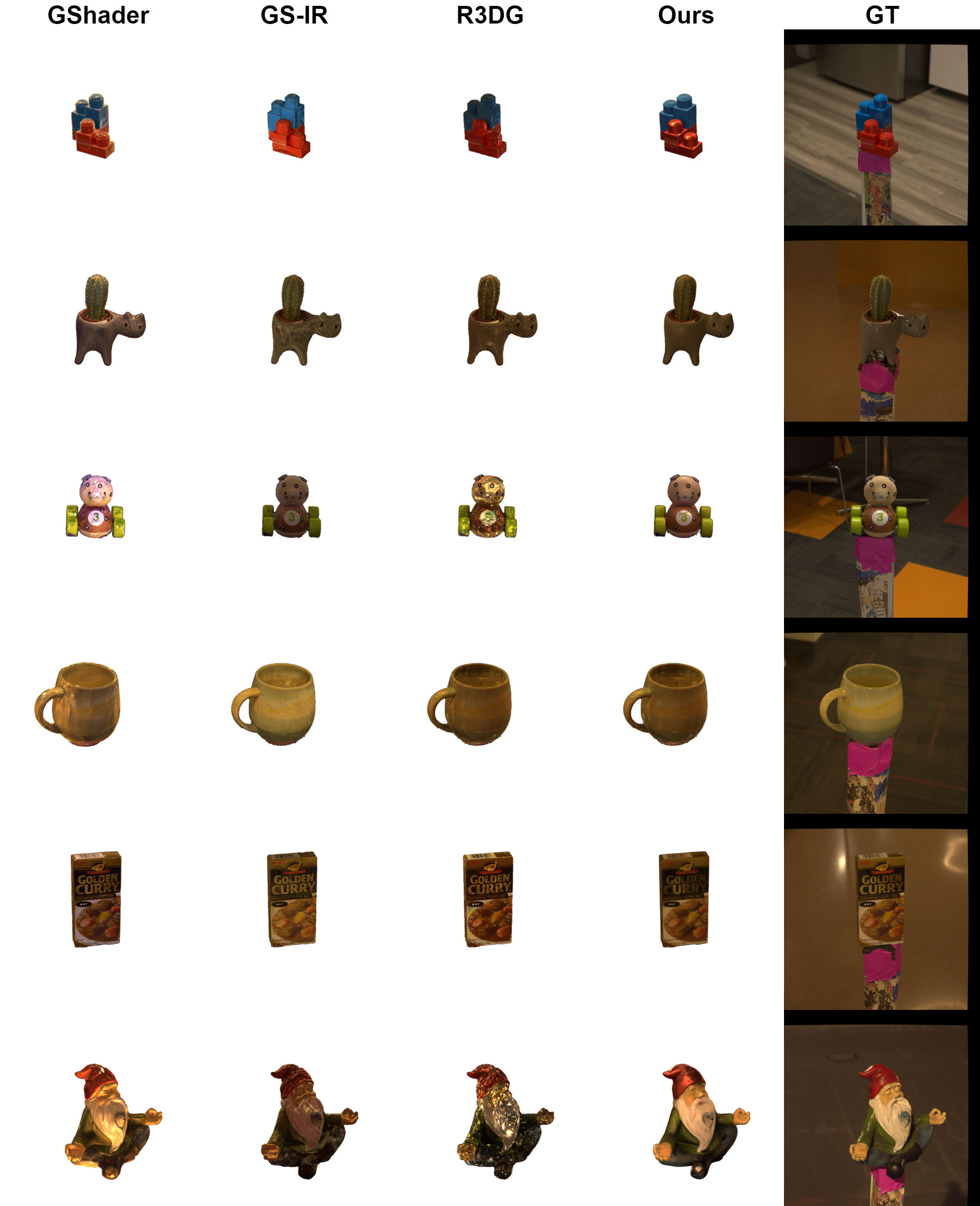}
    \caption{Relighting results on the StanfordORB dataset.}
    \label{fig:supp_exp_relighting_stanfordorb}
\end{figure*}

\paragraph{Material Decomposition and Relighting.}
% We show the PBR related material parameters decomposition in \cref{fig:supp_exp_decomposition_stanfordorb_a} and \cref{fig:supp_exp_decomposition_stanfordorb_b}. We render the corresponding parameter map by alpha blending according to \cref{eq:color_integration}. Comparing to other methods, our method can get a more reasonable decomposition of the PBR parameters with less noise. We also show the relighting results of different objects under several environment lights in \cref{fig:supp_exp_relighting_stanfordorb}.
\cref{fig:supp_exp_decomposition_stanfordorb} shows the PBR material decomposition in terms of albedo and roughness. From the results, our method can achieve a more reasonable decomposition of the PBR parameters while GS-IR tends to get high roughness and R3DG's roughness maps are noisy. We also show the relighting results in \cref{tab:supp_exp_relighting_stanfordorb} and \cref{fig:supp_exp_relighting_stanfordorb} to demonstrate the relighting quality further.

\begin{table*}[ht]
    \centering
    \begin{tabular}{l| cccc}
        \toprule[1pt]
        Methods & GShader        & GS-IR          & R3DG           & Ours           \\
        \midrule
        Baking  & 23.86 / 0.9586 & 23.99 / 0.9611 & 24.71 / 0.9638 & 24.16 / 0.9593 \\
        Ball    & 22.04 / 0.9219 & 23.48 / 0.9208 & 23.19 / 0.9156 & 25.17 / 0.9357 \\
        Blocks  & 27.09 / 0.9716 & 28.78 / 0.9707 & 27.94 / 0.9692 & 32.16 / 0.9812 \\
        Cactus  & 26.99 / 0.9681 & 31.94 / 0.9724 & 30.15 / 0.9650 & 33.79 / 0.9819 \\
        Car     & 25.13 / 0.9671 & 26.71 / 0.9670 & 26.71 / 0.9662 & 30.57 / 0.9802 \\
        Chips   & 29.94 / 0.9742 & 28.19 / 0.9673 & 28.29 / 0.9694 & 28.36 / 0.9696 \\
        Cup     & 25.90 / 0.9641 & 26.96 / 0.9592 & 25.59 / 0.9482 & 28.93 / 0.9709 \\
        Curry   & 27.37 / 0.9704 & 30.58 / 0.9666 & 29.76 / 0.9674 & 31.77 / 0.9702 \\
        Gnome   & 28.88 / 0.9502 & 27.92 / 0.9388 & 26.94 / 0.9254 & 31.04 / 0.9540 \\
        Grogu   & 25.05 / 0.9709 & 27.17 / 0.9684 & 25.17 / 0.9606 & 25.80 / 0.9737 \\
        Pepsi   & 22.24 / 0.9517 & 24.46 / 0.9533 & 22.82 / 0.9490 & 24.10 / 0.9608 \\
        Pitcher & 25.55 / 0.9525 & 27.43 / 0.9545 & 29.00 / 0.9530 & 29.53 / 0.9659 \\
        Salt    & 25.64 / 0.9616 & 23.61 / 0.9289 & 24.96 / 0.9524 & 24.74 / 0.9446 \\
        Teapot  & 24.58 / 0.9655 & 23.86 / 0.9588 & 23.99 / 0.9575 & 25.86 / 0.9708 \\
        \midrule
        Average & 25.73 / 0.9606 & 26.79 / 0.9563 & 26.37 / 0.9545 & 28.28 / 0.9656 \\
        \bottomrule[1pt]
    \end{tabular}
    \caption{The relighting quality in terms of PSNR\(\uparrow\) and SSIM\(\downarrow\) in the StanfordORB dataset, we report the average metrics of two evaluation scenes. The comparison shows our method achieves the highest rendering quality.}
    \label{tab:supp_exp_relighting_stanfordorb}
\end{table*}

% \subsection{More ablation study}

% {
% 	\small
% 	\bibliographystyle{ieeenat_fullname}
% 	\bibliography{main}
% }


\end{document}